\documentclass[11pt]{report}
\usepackage{geometry, upennstyle} % You may add additional packages below this line. Overleaf will warn you if your packages conflict with the template packages. See Chapter 1 for more information.

\usepackage[nonamebreak,round]{natbib}
\usepackage{amsmath}
\usepackage{amssymb}
\usepackage{setspace}
\usepackage{mathtools}
\usepackage[utf8]{inputenc}
\usepackage[T1]{fontenc}
\usepackage[thinc]{esdiff}

\setcitestyle{square}

\usepackage{xcolor}
\usepackage{xparse}
\usepackage{algorithm}
\usepackage{algpseudocode}
\usepackage{graphicx} %package to manage images
\graphicspath{ {images/} }
\NewDocumentCommand{\codeword}{v}{%
\texttt{\textcolor{blue}{#1}}%
}

\definecolor{codegreen}{rgb}{0,0.6,0}
\definecolor{codegray}{rgb}{0.5,0.5,0.5}
\definecolor{codepurple}{rgb}{0.58,0,0.82}
\definecolor{backcolour}{rgb}{0.95,0.95,0.92}
\usepackage{listings}

\lstdefinestyle{mystyle}{
    backgroundcolor=\color{backcolour},   
    commentstyle=\color{codegreen},
    keywordstyle=\color{magenta},
    numberstyle=\tiny\color{codegray},
    stringstyle=\color{codepurple},
    basicstyle=\ttfamily\footnotesize,
    breakatwhitespace=false,         
    breaklines=true,                 
    captionpos=b,                    
    keepspaces=true,                 
    numbers=left,                    
    numbersep=5pt,                  
    showspaces=false,                
    showstringspaces=false,
    showtabs=false,                  
    tabsize=2,
    linewidth=16cm,
    xleftmargin=10pt
}

\title{LATENT SPACE INFERENCE FOR SPATIAL TRANSCRIPTOMICS}
\author{\href{mailto:zsayem30@student.ubc.ca}{\normalsize Sayem Nazmuz Zaman} \normalsize (68943182), \href{mailto:david311@student.ubc.ca}{\normalsize David Wang} \normalsize (99030033), and
\href{mailto:poyuchen@alumni.ubc.ca}{\normalsize Daniel Chen} \normalsize (43419712)}

\gradgroup{2255 - Spatial Transcriptomics} % See https://provost.upenn.edu/phd-graduate-groups for names
\date{2023} % Enter four-digit year of deposit
\supervisor{Professor Jiarui Ding}
\supervisortitle{Assistant Professor at UBC Computer Science Department}

\gradchair{Graduate Group Chairperson's Typed Name, Full Faculty Title}
\committee{Dylan Gunn, Project Lab Director}
\committee{Miti Isbasescu, Head of Software Systems}
\committee{Bernhard Zender, Student Support} % If applicable, comment out or duplicate this line to remove or add a committee member
\begin{document}
\maketitle % If you are in the Romance Languages or the Managerial Science and Applied Economics graduate group, comment out this line. If you do, you will also need to edit lines 33 and 72.
%\makespecializationtitle % If in the Romance Languages or the Managerial Science and Applied Economics graduate group, uncomment this line. If you do, you will also need to edit lines 33 and 71.
\setcounter{page}{2}

%%%% OPTIONAL COPYRIGHT NOTICE
%\makecopyright % If not applicable, comment out this line to hide the optional traditional copyright notice page. If you do, you will also need to edit line 47.
%-------------------------------------------
%\makecreativecommons % If applicable, uncomment this line to insert the optional Creative Commons License copyright notice page. If you do, you will also need to edit lines 49, 51, and 76.
%-------------------------------------------

%%%% OPTIONAL DEDICATION PAGE
%\makededication % If not applicable, comment out this line to hide the optional dedication page. If you do, you will also need to edit line 55.
%-------------------------------------------

%%%% OPTIONAL ACKNOWLEDGEMENT PAGE
\makeacknowledgement % If not applicable, comment out this line to hide the optional acknowledgment page. If you do, you will also need to edit line 59.
%-------------------------------------------
In order to understand the complexities of cellular biology, researchers are interested in two important metrics: the genetic expression information of cells and their spatial coordinates within a tissue sample. However, state-of-the art methods, namely single-cell RNA sequencing and image based spatial transcriptomics can only recover a subset of this information, either full genetic expression with loss of spatial information, or spatial information with loss of resolution in sequencing data.

In this project, we investigate a probabilistic machine learning method to obtain the full genetic expression information for tissues samples while also preserving their spatial coordinates. This is done through mapping both datasets to a joint latent space representation with the use of variational machine learning methods. From here, the full genetic and spatial information can be decoded and to give us greater insights on the understanding of cellular processes and pathways. 

All results and code for this project is documented at our github repository: \\ https://github.com/2255-Spatial-Transcriptomics

\pagebreak
\tableofcontents
%%%% OPTIONAL LIST OF TABLES
\clearpage \phantomsection \addcontentsline{toc}{chapter}{LIST OF TABLES} \listoftables % If not applicable, comment out this line to hide the optional List of Tables
%-------------------------------------------

%%%% OPTIONAL LIST OF ILLUSTRATIONS
\clearpage \phantomsection \addcontentsline{toc}{chapter}{LIST OF FIGURES} \listoffigures % If not applicable, comment out this line to hide the optional List of Illustrations
%-------------------------------------------

%%%% OPTIONAL PREFACE
%\makepreface % If applicable, uncomment this line to insert the optional preface. If you do, you will also need to edit line 67.
%-------------------------------------------

%%%%% MAIN TEXT %%%%%
\begin{mainf}

\chpt{Commonly used Terms}
\begin{center}
\begin{table}[ht]
        \centering
        \scalebox{1}{
        \begin{tabular}{l l}
        \hline
         \textbf{Term} & \textbf{Description}\\\\
        \hline
        VAE & Variational Autoencoder \\\\
        VGAE & Variational Graph Autoencoder \\\\
        sc-RNA & Single-cell RNA sequencing (method) \\\\
        ibST & Image-based Spatial Transcriptomics (method) \\\\
        Single-cell RNA-sequencing dataset (scRNA) & $X_{sc}$ \\\\
        Image-based spatial transcriptomics dataset (ST) & $X_{st}$ \\\\
        Latent Representation of $X_{sc}$ & $Z_{sc}$ \\\\
        Latent Representation of $X_{st}$ & $Z_{st}$ \\\\
        Reconstructed data from $X$ using VAE/VGAE & $\widetilde{X}$ \\\\
        Highly Variable Genes & HVG \\\\
        \end{tabular}}
        \caption{Commonly Used Terms (Also See Table~\ref{tab:commonterms})}
    \end{table}
\end{center}

\chpt{Introduction}
\section{Information about the Sponsor}
Professor Jiarui Ding is an assistant professor at the Computer Science Department at the University of British Columbia. He was a postdoctoral associate at the Broad Institute of MIT and Harvard. His research interest mainly lies in the interception between probabilistic machine learning, visualization, and bioinformatics. He is experienced in various machine learning theories and approaches that could be applied to genetic sequencing data. In particular, he aims to understand how variational autoencoders can bridge the advantages between different RNA sequencing methods, which generates a novel approach to analyzing commercially available gene expression data. 

\section{Background and Significance of the Project}

A better understanding of cell and tissue biology can provide powerful insights into treating diseases and cancer. State-of-the-art research focuses on two main methods to dive deep into the genomics information contained within cells and tissues. Single-cell RNA sequencing (scRNA-seq) technologies measure the mRNA expression information at a single-cellular resolution. However, this technology involves separating cells from the tissue into individual cells for measurement, which results in the loss of spatial information about the cells' location within the tissue microenvironment. Retaining this spatial information can provide powerful insights into tissue architecture and cellular pathways. Image-based spatial transcriptomics can preserve the spatial information of sequenced genes. However, this method only allows the profiling of a smaller set of predefined genes (e.g. 500 compared to all 20,000 human genes) and at coarser resolutions.

Our project seeks to investigate a methodology for retaining both sets of information (full gene expression and spatial coordinates) when presented with only gene expression information by imputing the missing information through a probabilistic machine learning framework. This pipeline can provide a more comprehensive understanding of inter-cellular interactions utilizing gene expression and spatial information. 

Note that more advanced methods, such as MERFISH, can preserve spatial information while sequencing at an exceptionally high resolution. However, one major limitation of these methods is their relatively high cost, making them difficult for large-scale studies. Our project aims to maximize the use of existing data collection methods and provide an alternative analysis pipeline to those more advanced methods. 

\section{Project Objectives}
Our research aims to create a probabilistic machine learning model to associate gene expression data with spatial information. This would enable us to estimate the spatial position of cells based on the genetic information of RNA molecules in each cell. We will use two datasets - a single-cell RNA-sequencing dataset ($X_{sc}$) and an image-based spatial transcriptomics dataset ($X_{st}$) to train this statistical model and generate a shared latent space. This will facilitate the 'decoding' of the latent code of a cell from image-based spatial transcriptomics and allow us to obtain a comprehensive transcriptomic-wide measurement or fill in any unmeasured genes in image-based spatial transcriptomics. 

\section{Scope and Limitations}

\subsection{Scope}

The input of our project will be the scRNA sequencing data and the spatial transcriptomics data. The output of our project will be the inferred spatial location for a specific cell. 

In this project, we use existing VAE and VGAE algorithms that are readily available in the literature and have a working codebase. We slightly modified those algorithms to make them fit into our pipeline, but we are not reinventing the wheel. We also use existing data collected by research institutes such as 10x Genomics and Allen Brain Institute, which are processed cell-gene or spot-gene matrices instead of raw FASTQ files. 

\subsection{Limitations}

\subsubsection{Morphological differences between tissues of different patients}
The primary limitation of the project is that the two datasets ($X_{sc}$ and $X_{st}$) come from the same type of tissue type but might be from different patients. Although our model might infer spatial information of genes, there might be morphological differences between the tissues of different patients, so the spatial information inferred might need to be more accurate.

\subsubsection{Our Model does not generalize to unknown cell types}
If there are cell types which our model does not observe, then our model might fail when inferring the spatial location of these cell types. This might be very common in certain types of cancers where tumour populations are unique to a patient and might not have been observed before. 

\subsubsection{VAE assume a multidimensional normal prior for low-dimensional latent variables}
Our VAE model, scVI, maps high-dimensional data into low-dimensional Cartesian latent space, which causes points to be "squeezed" closer to one another, resulting in an inaccurate latent representation of the data. Suppose a hyperbolic latent space distribution was used with uniform prior distributions on a hypersphere, or a hyperbolic latent space representation was used with a wrapped normal distribution in hyperbolic space as the prior. In that case, the latent representation might have been more representative based on the results of \cite{scphere}. 

\subsubsection{VAE uses cosine distances to approximate proximity of cells when embedding data on a hypersphere}
Based on the results of \cite{scphere}, a hypersphere latent space representation seems to represent scRNA data better. Our VAE, scVI, uses cosine distance to measure the proximity between two cells when embedding in a lower dimensional space and embedding data on a hypersphere in Euclidean space distorts dimensional reduction. 

\subsubsection{VAE does not use batch correction}
In addition to this, single-cell profiles are impacted highly by diverse biological factors such as age, sex, diseases and batch effects in different experiments with different lab protocols. These batch corrections are handled outside of our VAE, scVI.

\chpt{Discussion} \label{discussion}
\section{Approach and System Overview}
The project aims to develop a probabilistic machine learning model to map the gene expression data to spatial information. However, associating high-dimensional gene expression data with another high-dimensional spatial transcriptomics data is significantly challenging. It is a common practice to reduce the dataset to its lower dimensional latent representation for further analysis. Variational autoencoders (VAE) have been proven effective in obtaining the optimal latent space for high-dimensional RNA sequencing data. On the other hand, we can use Variational graph autoencoders (VGAE) to obtain the latent spaces for spatial transcriptomics data \citep{sedr}. We will develop a probabilistic machine learning pipeline, called the "latent mapping pipeline," that consists of VAEs and VGAEs to associate those latent spaces with their original data. The latent mapping pipeline consists of three steps, and their simplified descriptions are as follows: 

\begin{enumerate}
    \item Obtain optimal latent space representation for scRNA dataset (Section~\ref{sec:step1})
    \item Map the gene expressions in the scRNA dataset to the gene expressions in the spatial transcriptomics dataset (Section~\ref{sec:step2})
    \item Map the gene expressions to the spatial information in the spatial transcriptomics dataset (Section~\ref{sec:step3})
\end{enumerate}

The in-depth explanation of the three steps are explained in the \nameref{sec:projectpipeline} section. Once this pipeline is trained on datasets containing specific cell types, we will be able to achieve our project goal: to impute the spatial information of a specific gene from the scRNA dataset, or vice versa. 

\section{Data Collection} \label{sec:datacollection}
We have two main techniques to study gene expression patterns in cells: \nameref{sec:sc} and \nameref{sec:st}. These two methods rely on RNA sequencing to capture and analyze gene expression data. Both techniques have become increasingly popular in the field of transcriptomics and are often used together to provide a more comprehensive understanding of gene expression in tissues. However, there are respective drawbacks for each method due to technical limitations. 

\subsection{Single Cell RNA Sequencing} \label{sec:sc}
Single-cell RNA sequencing (scRNA-seq) is a technique that enables researchers to study gene expression in individual cells. The procedure involves isolating single cells from a sample and capturing the RNA molecules in each cell. The RNA is then converted into DNA sequenced using a high-throughput method shown in Figure~\ref{fig:data-scrna}. We can analyze the resulting data and identify which genes are expressed in each cell to gain insights into how different cells function and interact. 

\begin{figure}[H]
    \centering
    \includegraphics[scale=0.4]{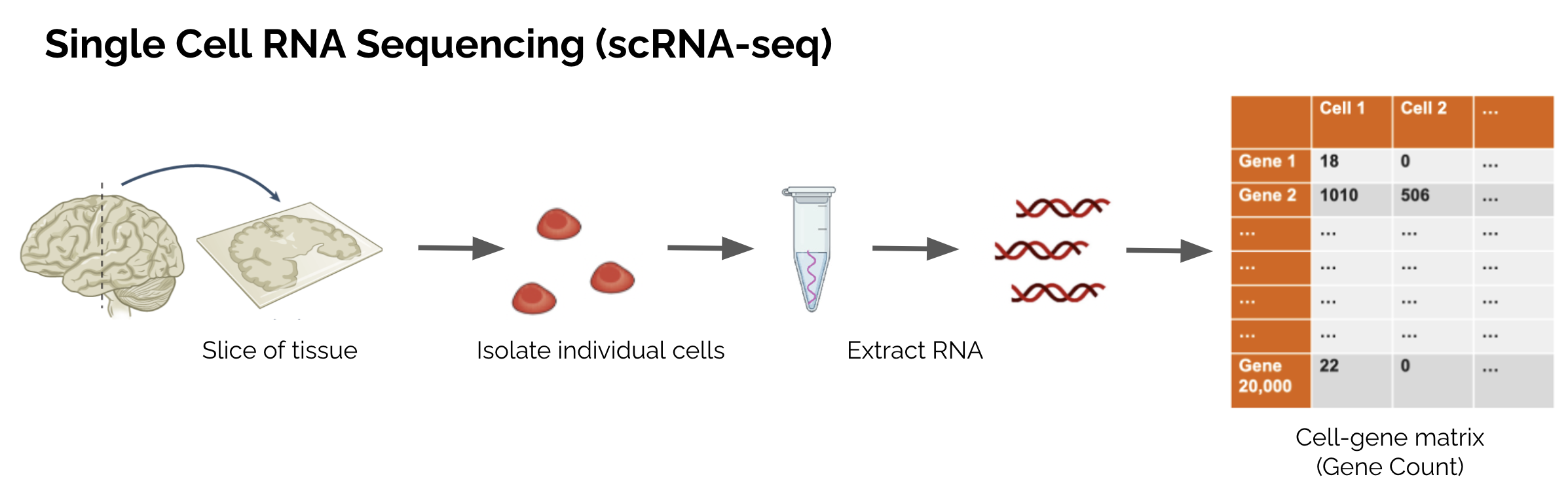}
    \caption{Data collection pipeline for scRNA sequencing}
    \label{fig:data-scrna}
\end{figure}

We use public datasets from companies and research institutions, such as 10x Genomics and Allen Institute for Brain Science. These datasets are already processed and are in the cell-gene 2D matrix form that contains the gene count for a specific gene in a specific cell type. Typically, the rows represent the gene types, and the columns represent the individual cells sequenced. 

\subsubsection{Data Processing}
It is worth noting that the data is often subject to different sources of noise that can affect its reliability and usability. One noise source is biological variability, which arises from differences in gene expression levels between individual cells or tissues due to genetic, epigenetic, or environmental factors. Another noise source is technical variability, which arises from errors introduced during the experimental and sequencing steps, such as PCR amplification bias, sequencing errors, or batch effects. Moreover, housekeeping genes are highly expressed across many different cell types, but their expression levels may not necessarily reflect the cells' actual biological functions or traits. Therefore, we must ensure the data is reliable and perform preliminary data processing to remove housekeeping genes. The detailed data processing procedure is described in \nameref{sec:dataprocessing} section.

The processed cell-gene matrix is denoted as $\boldsymbol{X_{sc}}$, which will be the input datasets for our pipeline. 

\subsubsection{Drawbacks for scRNA Sequencing Data}
The main drawback of scRNA sequencing data is that the spatial information of the genes would be lost while isolating individual cells from a tissue sample. This procedure could damage some cells and prevent us from knowing where the genes were located prior to the sequencing process.

\subsection{Spatial Transcriptomics}
\label{sec:st}
Spatial transcriptomics is a technique that enables researchers to study gene expression in cells within their natural tissue context. The procedure involves collecting RNA molecules from a tissue sample using a unique slide with probes designed to detect specific RNA molecules. The RNA molecules are then converted into cDNA and sequenced, allowing researchers to map the spatial distribution of gene expression patterns in the tissue sample. This technique provides valuable information about how different cells are arranged and interact with each other in their natural environment, which is essential for understanding complex biological systems, cellular pathways and evolutionary trajectories. 

\begin{figure}[H]
    \centering
    \includegraphics[scale=0.4]{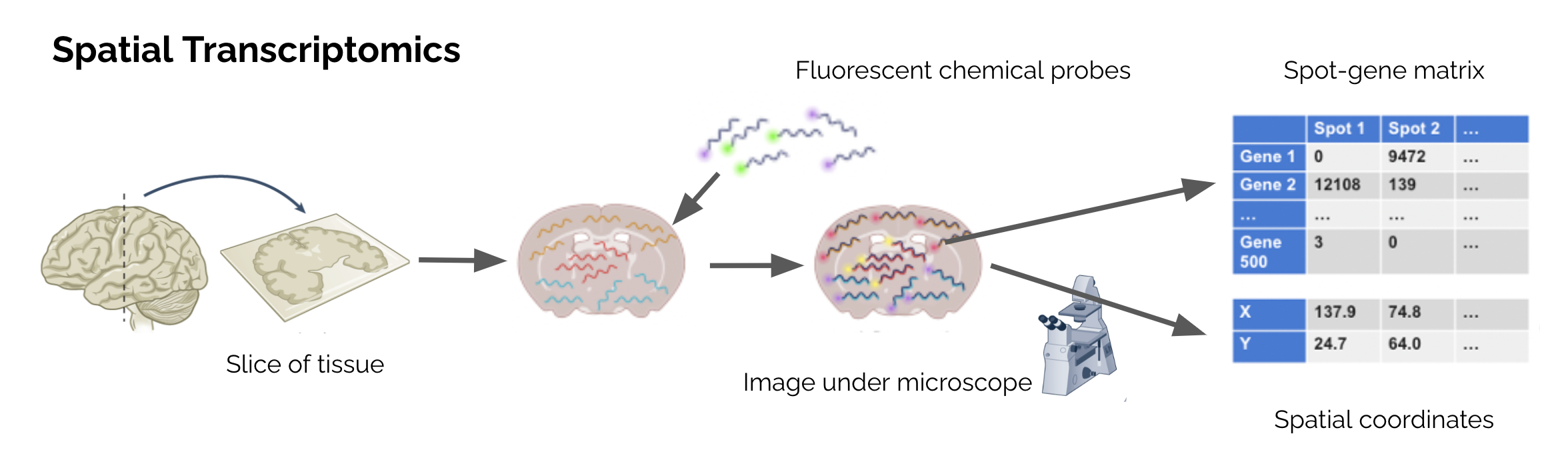}
    \caption{Data collection pipeline for spatial transcriptomics}
    \label{fig:data-spatial}
\end{figure}

We also gathered publicly-released spatial transcriptomics datasets that contain similar cell types as $\boldsymbol{X_{sc}}$. The dataset consists of two components: the spots' spatial location ($x$, $y$ coordinates) and the spot-gene matrix. The latter is a 2D matrix representing the gene expression levels for each spot on a tissue slice. "Spot" refers to a defined area on the slice where RNA molecules have been captured and sequenced. Similar to analyzing scRNA sequencing data, we perform data pre-processing on the raw spatial transcriptomics data described in \nameref{sec:dataprocessing} section and obtain $\boldsymbol{X_{st}}$. 

\subsubsection{Drawbacks for Spatial Transcriptomics Data}
Although we can obtain the spatial information of the spots in spatial transcriptomics, we can only profile a small number of genes per spot due to chemical limits (number of chemical probes per assay and finite RNA molecule capturing efficiency for each probe). Generally speaking, we can only capture around 500 genes in spatial transcriptomics.

\subsection{Challenges in Associating Two Datasets}

Now, we have two data collection methods that produce the datasets we are working with. However, associating high dimensional gene expression data to another high dimensional spatial transcriptomics data is significantly challenging due to a number of reasons. First, the noise or variability in the data can obscure core patterns or features, making it difficult to distinguish between random fluctuations and true biological differences. Secondly, high-dimensional datasets are often noisy, complex and nonlinear, making it challenging to visualize or interpret the data in a meaningful way. Last but not least, they are also confounded by technical factors such as batch effects or differences in experimental conditions. The simplest and most common way to solve these issues is to obtain each high-dimensional dataset's low-dimensional latent representation using VAEs and VGAEs and perform further analysis on them.

\section{Selecting Machine Learning Frameworks: VAE and VGAE}

\subsection{Benchmarking optimal VAEs}
\subsubsection{Overview}

\begin{figure}[H]
    \centering
    \includegraphics[scale=0.4]{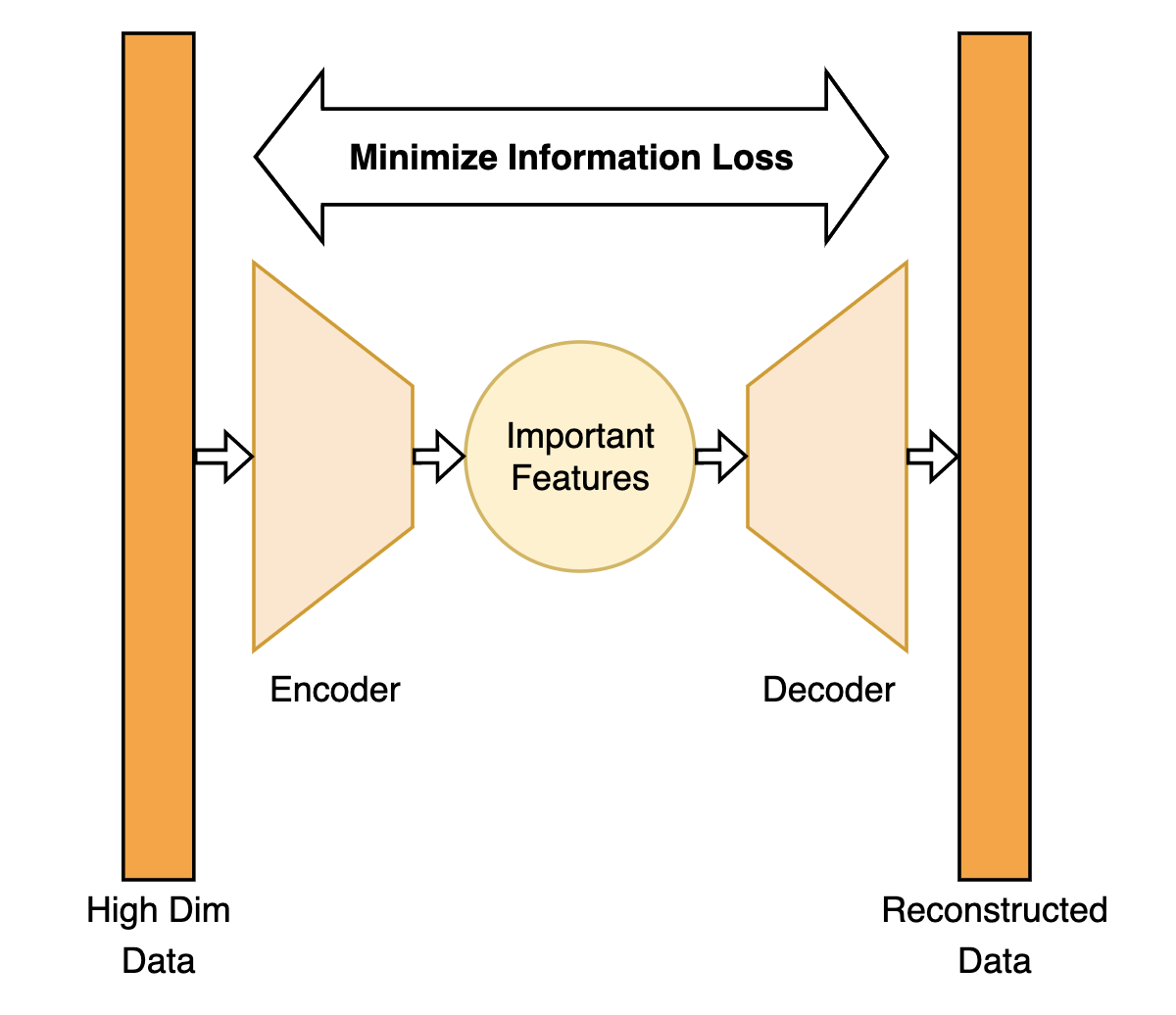}
    \caption{\label{fig:vae}Simple Architecture of a Variational Autoencoder}
\end{figure}

We utilize Variational Autoencoders (VAE) to obtain the optimal latent space that accurately represents the high dimensional sc-RNA data ($X_{sc}$). Traditional VAEs is a neural network that contains an encoder and a decoder. The encoder produces the latent representation of the input data, while the decoder tries to reconstructs the latent representation back to the original data. The encoder and decoder network jointly minimizes the loss between the original data and the reconstructed data, as shown in Figure~\ref{fig:vae}. Many VAE architectures have been implemented in literature and to not reinvent the wheel, we focused on benchmarking five VAE methods that were designed for analyzing single-cell RNA sequencing data: 

\begin{enumerate}
    \item \nameref{scphere_overview} \newline \citep{scphere}
    \item \nameref{scDHA_overview} \citep{Tran799817}
    \item \nameref{VASC_overview} \citep{wang_gu_2018}
    \item \nameref{DRA_overview} \citep{lin_mukherjee_kannan_2020}
    \item \nameref{scVI_overview} \citep{scVI}
\end{enumerate}

All of these methods remove gene data with an insignificant contribution to the core features that represent the dataset and minimizes the loss when reconstructing the dataset from the latent features. To analyze the performance of each method, we extracted the latent representations from the same set of data using these methods. Then, we select the best performing VAE by evaluating the quality of their corresponding latent spaces (see \nameref{sec:overview_benchmark}). 

\begin{figure}[H]
    \centering
    \includegraphics[scale=0.6]{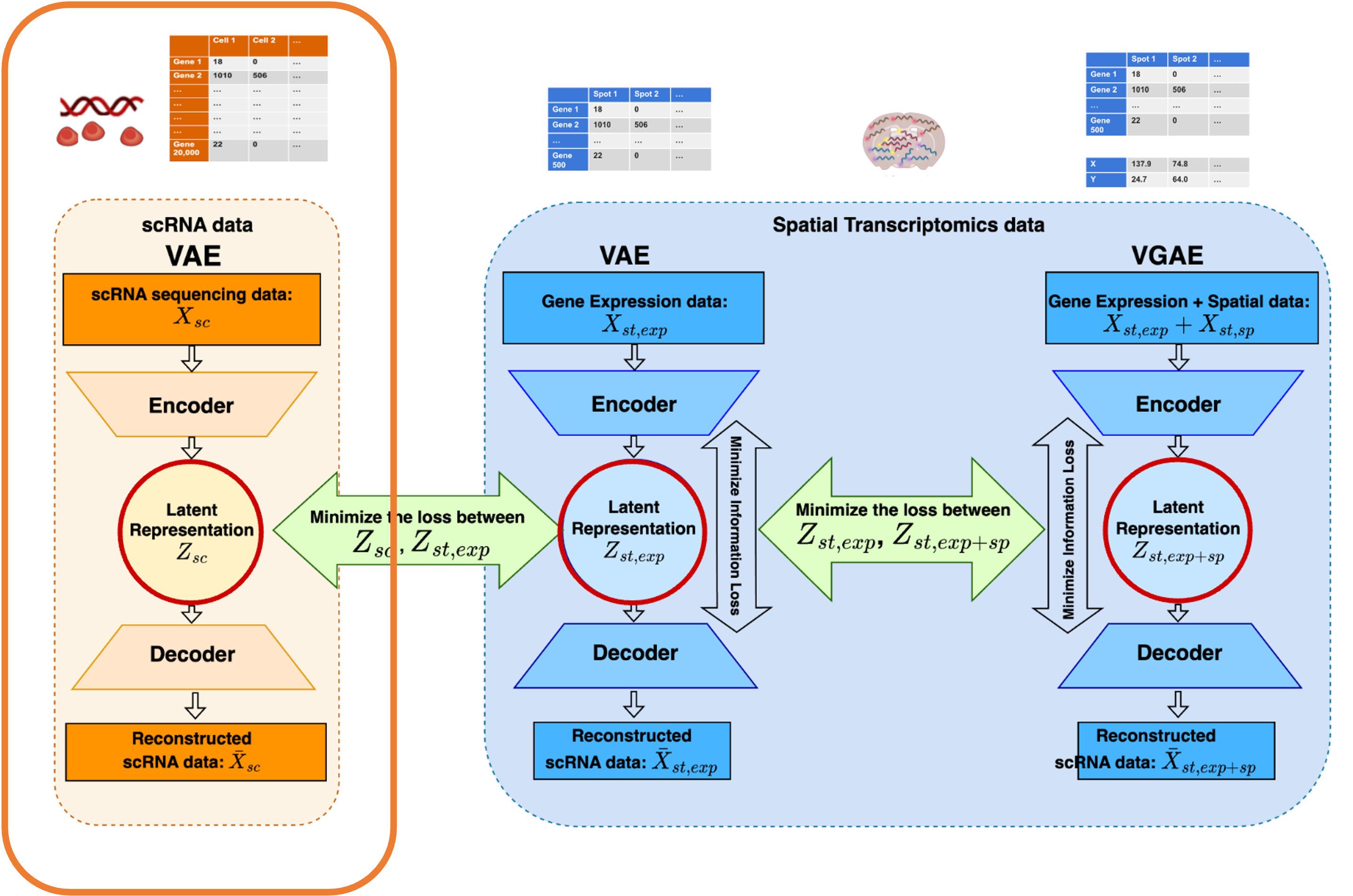}
    \caption{The red circle indicates the VAE portion of the pipeline, which is the purpose of this benchmarking process.}
\end{figure}

\subsubsection{Benchmark Approach and Evaluation Metrics} 
\label{sec:overview_benchmark}
There are numerous metrics to evaluate a VAE's performance, including reconstruction loss, performance in downstream tasks (generation and classification), and the evidence lower bound (ELBO). We performed \textbf{K-nearest neighbours (kNN) with k-fold cross validation} on the latent embeddings, then compared those embeddings of the nearest neighbours to the original input data to evaluate the quality of the latent space. We used a $75/25$ split for the training/validation dataset and calculated the final accuracy using the following formula

$$ accuracy(y, \hat{y}) = \frac{1}{n_{\text{samples}}} \sum_{i=1}^{n_{\text{samples}}} \mathbb{I}^{\hat{y}_i = y_i} \text{, where } \mathbb{I}^{\hat{y}_i = y_i} \text{ is $1$ if $\hat{y}_i = y_i$ and $0$ otherwise}$$

There are a few key points to consider while using kNN to evaluate the performance of VAEs: 

\begin{enumerate}
    \item \textbf{Quality of Input Data: } We need to ensure that the input data is not overly noisy, does not contain an unreasonable amount of outliers, and has a roughly equal distribution of different cell types for kNN evaluation to be accurate. In our analysis, we use \hyperref[datasets]{datasets} provided by the Broad Institute of MIT and Harvard.
    \item \textbf{Number of $k$: } We need a appropriate number of nearest neighbours $k$ to obtain an accurate evaluation of the model. Generally, the model obtains the highest accuracy at around $k = \{\text{number of cell types}\}$. The higher the $k$, the smoother the classification boundary, and the lower the misclassification loss, but the more computationally expensive. In our analysis, we vary $k$ to obtain the latent representation that can most accurately express the initial data distribution. 
    \item \textbf{Choice of distance metric: } 
    We chose to use the \textbf{Euclidean distance} for our kNN algorithm for our project.
    $$ \textrm{dist}(x, y) = \sqrt{\sum_{i=1}^n(x_i - y_i)^2} $$
\end{enumerate}

\subsubsection{K-Fold Cross Validation}

In our analysis, 4-fold cross-validation was used: each dataset was split into 4 groups, and the model was trained and tested 4 separate times, with each group being the test set during different iterations. This ensures uniformity when training and testing the data as we ensure that our model is not only sensitive to a subset of the data and we are interested in observing how our model performs regardless of the data subset being used to train and test.  We then compare the average model performance and accuracy along with the standard deviation across the 4 different folds.

\subsubsection{Datasets} \label{datasets}
All the datasets that were used were from the Single Cell Portal provided by \href{https://singlecell.broadinstitute.org/single_cell}{Broad Institute of MIT and Harvard}. We particularly worked with 3 main datasets:
\begin{itemize}
    \item \textbf{Dataset A}: lung\_human\_ASK440 \newline
    \underline{Description:} A sample tissue of the upper left lobe of the lung from a male donor aged 58 who was a smoker.
    \item \textbf{Dataset B}: Adipose \newline
    \underline{Description:} Loosely connective tissue that forms body fat.
    \item \textbf{Dataset C}: UC\_Stromal \newline
    \underline{Description:} This dataset consists of stromal cells from a Human umbilical cord.
\end{itemize}

See \nameref{dataset_dist} for the cell type distributions of the datasets.

\subsubsection{Source Code Arrangement}
The source code for the benchmarking pipeline is published on \href{https://github.com/2255-Spatial-Transcriptomics/benchmark}{GitHub}. The code base consists of 4 submodules that are forked from their respective repositories.

\subsection{Selecting optimal VGAE}
Similar to variational autoencoders, variational graph autoencoders are able to obtain the latent representation of a given high dimensional dataset. However, the encoder of the VGAE consists of  graph convolutional networks (GCNs) that is able to encode and decode graph-structured data, which is precisely useful for our spatial transcriptomics datasets. It takes in a adjacency matrix and a feature matrix, in our case, the spot-gene matrix, and outputs the learned latent representation of both input matrices.

The method SEDR (Unsupervised spatially embedded deep representation of spatial transcriptomics) developed a VGAE specifically for analyzing spatial transcriptomics data \citep{sedr}. We decided to make use of their work and incorporate their VGAE architecture into our latent mapping pipeline.

\begin{figure}[H]
    \centering
    \includegraphics[scale=0.6]{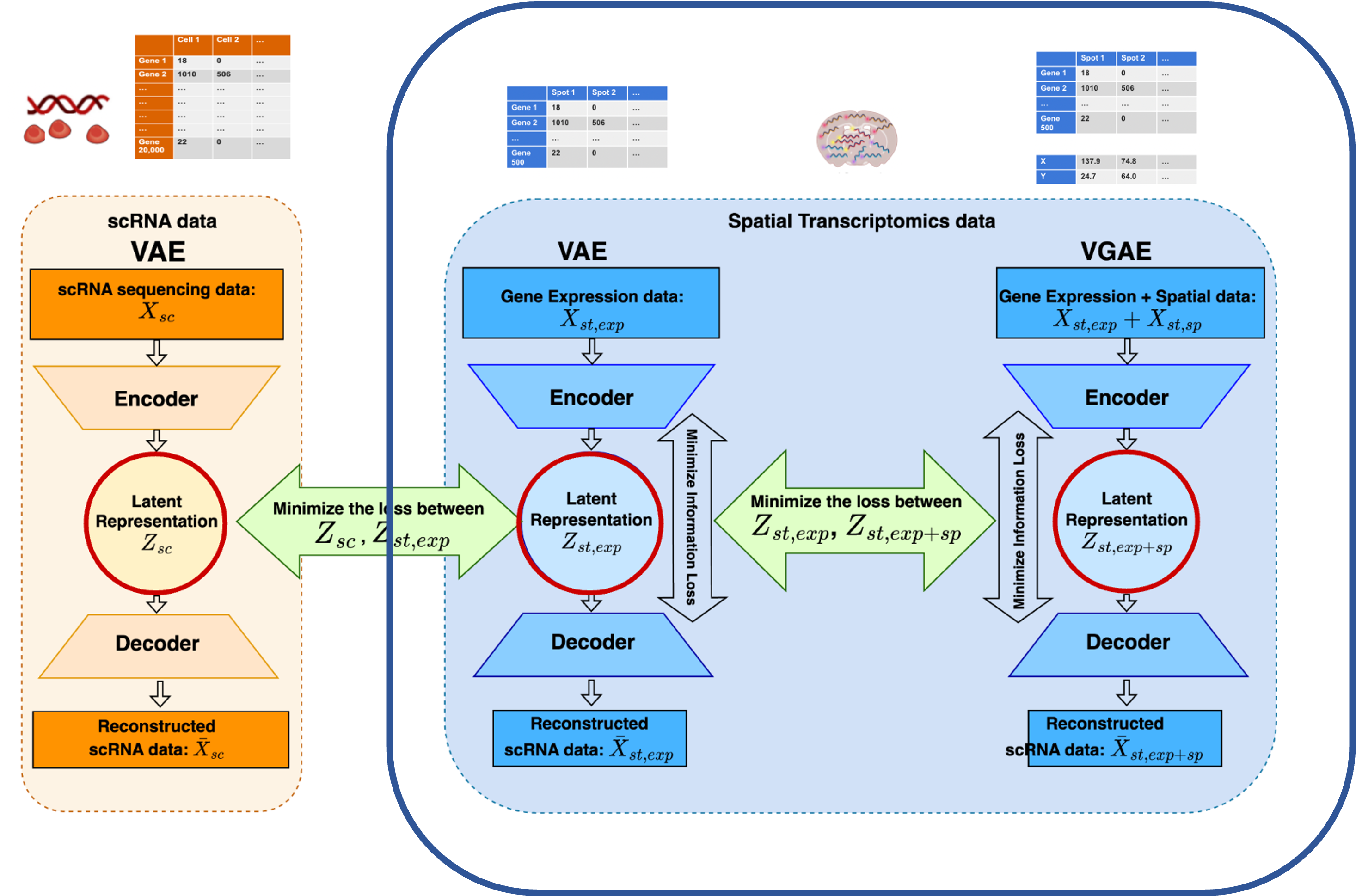}
    \caption{The blue circle indicates the VGAE portion of the pipeline, which inherits the architecture from SEDR \citep{sedr}.}
\end{figure}

\section{Theory and Model Functionality}
\subsection{Data Processing}\label{sec:dataprocessing}

Our first task with both the datasets is to apply some quality control measures. This helps to reduce the size of our datasets. We first remove cells which have less than 200 genes expressed and remove genes which are expressed in less than 60 cells. These numbers were suggested by our capstone supervisor and are an extrapolation of the values used for analyzing a dataset of Peripheral Blood Mononuclear Cells (PBMC)[\cite{Satija2015}]. 

Next, we remove cells with high mitochondrial and ribosomal genes. High proportion of these cells indicate loss of cytoplasmic RNA from perforated cells. 

Since we are given two extremely high-dimensional datasets, we would want to select only the genes which are relevant and have high probability of being mapped. These genes correspond to genes with high variability. The discovery of the highly variable gene (HVG), allows detection of genes that correspond to cell-to-cell variation within a homogeneous cell population. We have the underlying assumption that the highly variable genes are more likely to be mapped between the two datasets.

The top 2000 highly variable genes were selected from the scRNA dataset ($X_\text{{sc, exp}}^{2000}$). In addition to this, the top 500 highly variable genes shared between the scRNA dataset and the Spatial Transcriptomics dataset were also processed for later use ($X_\text{{sc, exp}}^{500}, X_\text{{st, exp + sp}}^{500}$). 

\begin{center}
\begin{table}[!h]
        \centering
        \scalebox{1}{
        \begin{tabular}{l l}
        \hline
         \textbf{Term} & \textbf{Description}\\
        \hline
        $X_{\textrm{sc, exp}}^n$ & \textrm{The cell gene matrix of single-cell RNA sequencing data, with n genes} \\\\
        $X_{\textrm{st, exp}}^n$ & \textrm{The spot gene matrix of spatial transcriptomics data, with n genes} \\\\
        $X_{\textrm{st, sp}}^n$ & \textrm{The spatial locations for spots for spatial transcriptomics data, with n genes} \\\\
        $X_{\textrm{st, exp+sp}}^n$ & \textrm{The gene expression and spatial locations for spots for spatial } \\& transcriptomics data, with n genes \\\\
        $Z_{i}^n$ & \textrm{The latent space of dataset i, for n genes} \\\\
        $\bar{Z}_{i}^n$ & \textrm{The finalized, \textbf{fixed} latent space of dataset i, for n genes} \\\\
        $\tilde{X}_i^n$ & \textrm{The reconstructed data for dataset i, with n genes} \\\\
        \end{tabular}}
        \caption{\label{tab:commonterms} Commonly Used Terms for Project Pipeline}
    \end{table}
\end{center}

\subsection{Project Pipeline} \label{sec:projectpipeline}

Since we have two datasets, scRNA dataset and Spatial Transcriptomics dataset containing two different types of information (scRNA data contains only gene expression data and Spatial Transcriptomics data contains both gene expression data and spatial information of the cells), it is difficult to map these two datasets together. We solve this problem by mapping the latent spaces generated by VAEs and VGAEs of the two datasets. We do this in three steps.

\subsubsection{Step 1: Find ideal Latent Space representation of the top 2000 highly variable genes of the scRNA dataset} \label{sec:step1}

In step 1, a VAE framework is implemented to generate an ideal, compact latent space representation for the 2000 HVG of the scRNA data. The encoder network of the VAE, $f_{\textrm{vae}}$, takes the original data and passes it through a series of neural networks to find a latent space representation: 
\begin{align*}
    Z_{\textrm{sc, exp}}^{2000} = f_{\textrm{vae1}}(X_{\textrm{sc, exp}}^{2000})
\end{align*}
The decoder network of the VAE, $f^{-1}_{\textrm{vae}}$, takes the latent space representation and tries to reconstruct the original data from the latent space representation:
\begin{align*}
    \widetilde{X}_{\textrm{sc, exp}}^{2000} = f^{-1}_{\textrm{vae1}}(Z_{\textrm{sc, exp}}^{2000})
\end{align*}
The ideal latent space is found by minimizing both the KL divergence between the latent space obtained and the original data and the reconstruction loss between the original data and the reconstructed data. This ideal latent space representation of the 2000 HVG of the scRNA data is then \textbf{fixed}, and we represent this as $\bar{Z}_{\textrm{sc, exp}}^{2000}$ 

\begin{figure}[H]
    \centering
    \includegraphics[scale=0.5]{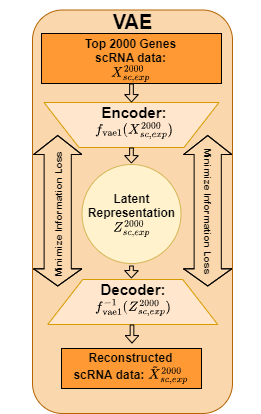}
    \caption{\textbf{Step 1} - Obtaining ideal latent space representation for 2000 Highly Variable genes of scRNA data}
    \label{fig:step1}
\end{figure}

\subsubsection{Step 2: Map the 500 shared genes between the 2000 HVG scRNA data and the Spatial Transcriptomics data} \label{sec:step2}

Once we have fixed the latent space representation of the 2000 HVG of the scRNA data, $\bar{Z}_{\textrm{sc, exp}}^{2000}$, we try to map the shared 500 genes between the 2000 HVG of the scRNA data and the spatial transcriptomics data. This is done in two sub steps:

\begin{enumerate}
    \item \textbf{Step 2.1:} We want to find the ideal Latent Space representation of the shared 500 genes of the scRNA data. This ideal representation is found by minimizing the KL divergence between the latent space and the original data (shared 500 genes of the scRNA data) and minimizing the loss between the original data and the reconstructed scRNA data of the shared 500 genes from its latent space representation. 
    \begin{align*}
        Z_{\textrm{sc, exp}}^{500} = f_{\textrm{vae1}}(X_{\textrm{sc, exp}}^{500}) \\
    \widetilde{X}_{\textrm{sc, exp}}^{500} = f^{-1}_{\textrm{vae1}}(Z_{\textrm{sc, exp}}^{500})
    \end{align*}
    
    In addition to this, to ensure that the latent space of the 500 genes of the scRNA data, $Z_{\textrm{sc, exp}}^{500}$, is \textbf{similar} to the fixed latent space representation of the 2000 HVG of the scRNA data, $\bar{Z}_{\textrm{sc, exp}}^{2000}$, from \textbf{Step 1} we implement a euclidean loss, $L_1$:
    $$L_1 = \textrm{dist}(Z_{\textrm{sc, exp}}^{500},  \bar{Z}_{\textrm{sc, exp}}^{2000})$$

    Minimising this loss ensures that the two latent spaces are similar to each other. 
    Once we have found an ideal latent space representation for the shared 500 genes for the scRNA data by minimizing KL divergence, reconstruction loss and euclidean loss between the latent space representation of the 500 shared genes and the 2000 HVG of the scRNA dataset, this latent space representation is then \textbf{fixed} and is represented by 
    $\bar{Z}_{\textrm{sc, exp}}^{500}$. 
    This is shown in Figure \ref{fig:step2_1}.
    \begin{figure}[H]
        \centering
        \includegraphics[scale=0.5]{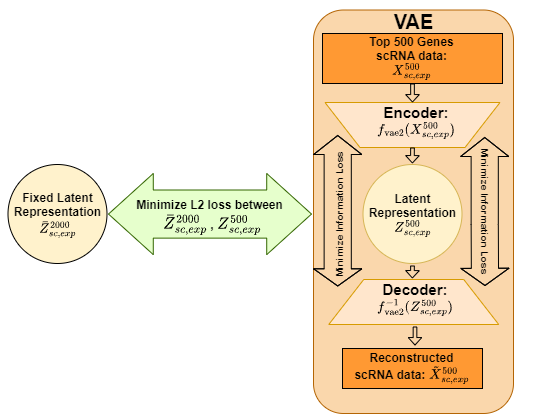}
        \caption{\textbf{Step 2.1} - Obtaining ideal latent space representation for 500 Highly Variable genes of scRNA data (shared with Spatial Transcriptomics) by reducing loss between $\bar{Z}_{\textrm{sc, exp}}^{2000}$ and $Z_{\textrm{sc, exp}}^{500}$}
        \label{fig:step2_1}
    \end{figure}    
    
    \item \textbf{Step 2.2:} We also want to find the ideal Latent Space representation of the shared 500 genes of the ST gene expression data. This ideal representation is found by minimizing the KL divergence between the latent space and the original data (shared 500 genes of the ST gene expression data), and minimizing the loss between the original data and the reconstructed ST gene expression data of the shared 500 genes from its latent space representation. Since the gene expression of the ST data does not contain any spatial information, we can use a VAE for this purpose:
        \begin{align*}
        Z_{\textrm{st, exp}}^{500} = f_{\textrm{vae2}}(X_{\textrm{st, exp}}^{500}) \\
        \widetilde{X}_{\textrm{st, exp}}^{500} = f^{-1}_{\textrm{vae2}}(Z_{\textrm{st, exp}}^{500})
    \end{align*}

    Now, since the ST gene expression data for 500 genes contain the same \textbf{type} (gene-expression) of information as the scRNA shared 500 genes, we can map their respective latent spaces together but in order to do so, we have to minimize an additional loss function. This loss function is adversarial instead of plain euclidean since we are trying make two latent spaces from different datasets (and thus different distributions) similar. Here we introduce a discriminator network $\textbf{D}$, which we will train to differentiate the two latent spaces: the \textbf{fixed} Latent space of the 500 shared genes for the scRNA data, $\bar{Z}_{\textrm{sc, exp}}^{500}$, and the latent space of the 500 shared genes for the ST gene expression data, $Z_{\textrm{st, exp}}^{500}$.
    
    This can help in finding the relationships between highly complex distributions. We define the loss function 
    \begin{align*}
        L_2 = \textbf{D}(\bar{Z}_{\textrm{sc, exp}}^{500}, Z_{\textrm{st, exp}}^{500})
    \end{align*}
    as the prediction accuracy the network has on differentiating between the labels of the respective latent spaces.

    We first train the discriminator to maximize $L_2$ so that our discriminator can correctly differentiate the two latent spaces and identify their labels correctly. Once the discriminator is trained, we then try to minimize $L_2$ when training the VAE, $f_{\textrm{vae2}}$, so that we generate a latent space, $Z_{\textrm{st, exp}}^{500}$, which our discriminator fails to differentiate from the \textbf{fixed} latent space, $\bar{Z}_{\textrm{sc, exp}}^{500}$. Once this loss is minimized, the latent space representation for the 500 shared genes of the ST gene data is \textbf{fixed},  $\bar{Z}_{\textrm{st, exp}}^{500}$
    
    All of this is shown in Figure \ref{fig:step2_2}.
    
    \begin{figure}[H]
        \centering
        \includegraphics[scale=0.5]{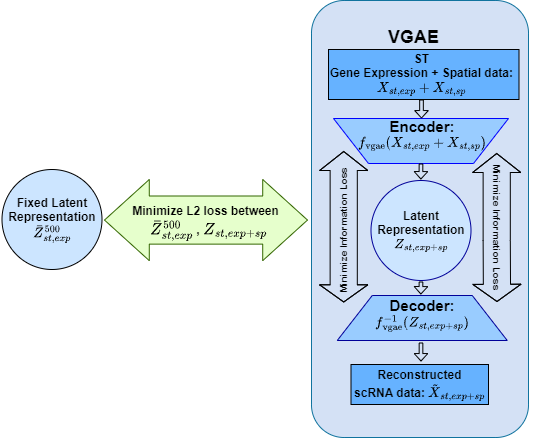}
       \caption{\textbf{Step 2.2} - Obtaining ideal latent space representation for 500 Highly Variable genes of ST gene expression data (shared with scRNA data) by reducing adversarial loss between $\bar{Z}_{\textrm{sc, exp}}^{500}$ and $\bar{Z}_{\textrm{sc, exp}}^{500}$}
        \label{fig:step2_2}
    \end{figure}    
    
\end{enumerate}

\subsubsection{Step 3: Map the ST Gene Expression latent space to the a generated latent space }\label{sec:step3}

Now that we have the \textbf{fixed} latent space representation for the 500 shared genes of the ST gene data is \textbf{fixed},  $\bar{Z}_{\textrm{st, exp}}^{500}$, we try to map this to the latent space representation of the full ST data containing both the gene expression and spatial information. In order to do so we first need to generate a latent space for the ST data and this latent space must encapsulate both the gene expression and spatial information into one complete latent space. 

To do so we introduce a Variational Graph Autoencoder (VGAE), which embeds both expression and spatial information into a graph neural network, and forms a latent space representation for the combined information. The encoder network and decoder network for a VGAE works in the same way as that for the VAE:
    \begin{align*}
        Z_{\textrm{st, exp+sp}}^{500} = f_{\textrm{vgae}}(X_{\textrm{st, exp+sp}}^{500}) \\
        \widetilde{X}_{\textrm{st, exp+sp}}^{500} = f^{-1}_{\textrm{vgae}}(Z_{\textrm{st, exp+sp}}^{500})    
    \end{align*}
    Just like the VAEs, the ideal representation is found by minimizing the KL divergence between the latent space and the original data (ST gene expression and spatial information) and minimizing the loss between the original data and the reconstructed ST data. 

    We also ensure that the latent space of the full ST data (gene expression and spatial information), $Z_{\textrm{st, exp+sp}}^{500}$, is \textbf{similar} to the fixed latent space representation of the 500 shared genes of the ST gene expression data obtained from \textbf{Step 2.2}, $\bar{Z}_{\textrm{st, exp+sp}}^{500}$. This can be done by implementing another euclidean loss, $L_3$ since both the latent spaces are from the same dataset thus have similar distributions:
    \begin{align*}
        L_3 = \textrm{dist}(Z_{\textrm{st, exp+sp}}^{500},\bar{Z}_{\textrm{st, exp}}^{500})
    \end{align*}

    Minimising this loss ensures that the two latent spaces are similar to each other. Step 3 is shown in Figure \ref{fig:step3}
    
    \begin{figure}[H]
        \centering
        \includegraphics[scale=0.5]{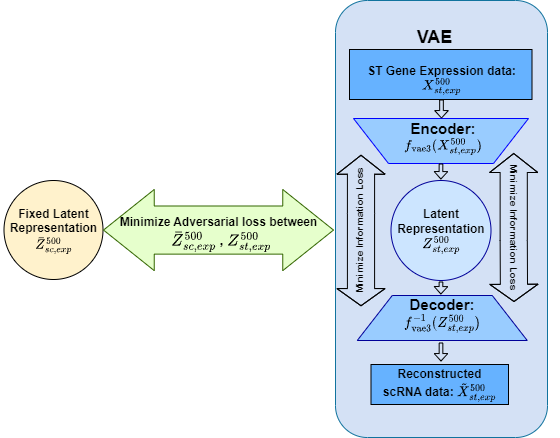}
       \caption{\textbf{Step 3} - Obtaining ideal latent space representation for full ST data} 
       \label{fig:step3}
    \end{figure}  
    
\section{Inference}
Once all our networks are trained, and losses have been minimized, we can perform inference to infer the spatial information of genes from scRNA dataset.

From \textbf{Step 2} and \textbf{Step 3}, we have mapped various latent spaces to each other by making their distributions similar. We can define these mappings as function, which takes a latent space, with some loss and generates the mapped latent space:

    \begin{align*}
        &\textrm{\textbf{Step 2.2:}} \quad \bar{Z}_{\textrm{st, exp}}^{500} \sim \bar{Z}_{\textrm{sc, exp}}^{500} \Rightarrow \bar{Z}_{\textrm{st, exp}}^{500} = f_{1}(\bar{Z}_{\textrm{sc, exp}}^{500}) \\
        &\textrm{\textbf{Step 3:}} \quad \bar{Z}_{\textrm{st, exp+sp}} \sim \bar{Z}_{\textrm{st, exp}}^{500} \Rightarrow \bar{Z}_{\textrm{st, exp+sp}} = f_{2}(\bar{Z}_{\textrm{st, exp}}^{500})
    \end{align*}
    Similarly, we have trained our VAE encoder network from \textbf{Step 2.1} and our VGAE decoder network from step 3:
    \begin{align*}
        &\textrm{\textbf{Step 2.1:}} \quad \bar{Z}_{\textrm{sc, exp}}^{500}= f_{\textrm{vae2}}(X_{\textrm{sc, exp}}^{500}) \\
        &\textrm{\textbf{Step 3:}} \quad \widetilde{X}_{\textrm{st, exp+sp}} = f^{-1}_{\textrm{vgae}}(X_{\textrm{st, exp+sp}}) \\
    \end{align*}

If we isolate the trained encoder network for the VAE in \textbf{Step 2.1}, the mapping functions for \textbf{Step 2.2}, \textbf{Step 3} and the trained decoder function from \textbf{Step 3}, we can build our inference model. This is shown in Figure \ref{fig:Inference}. 
    
    \begin{figure}[H]
        \centering
        \includegraphics[scale=0.45]{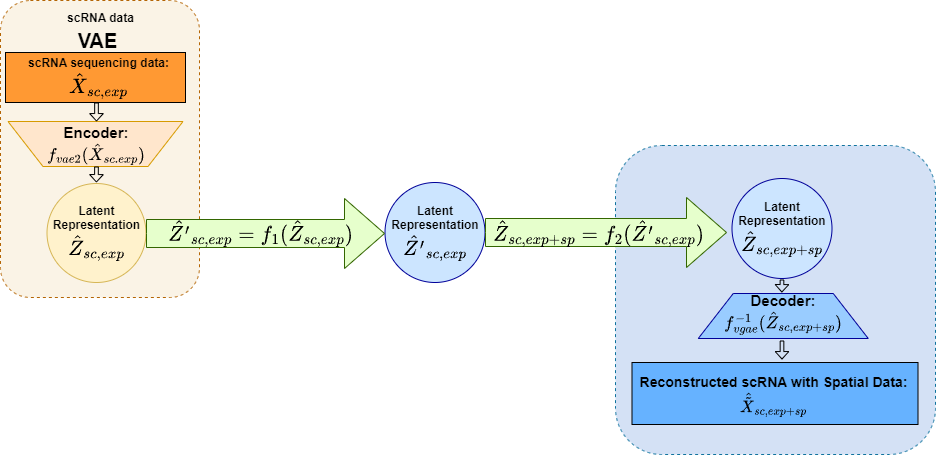}
       \caption{\textbf{Inference} - We can infer the spatial information of genes from scRNA dataset which do not have spatial information} 
       \label{fig:Inference}
    \end{figure}  

Following the inference pipeline in Figure \ref{fig:Inference}, we get:
\begin{align*}
    \hat{\bar{X}}_{\textrm{st, exp+sp}} = f^{-1}_{\textrm{vgae}}(f_2(f_1(f_{\textrm{vae2}}(X_{\textrm{sc, exp}})))) = f_{\textrm{model}}(X_{\textrm{sc, exp}})
\end{align*}

Thus, our inference model allows us to infer the spatial information of the gene expression data for the scRNA data.

\chpt{Implementation} \label{chpt:implementation}

Two popular libraries in machine learning are PyTorch and Tensorflow. After a preliminary analysis, our team decided to utilize PyTorch for the implementation of our models due to its ease of modification and pythonic implementation style. Based on prior experience and sponsor recommendations, we also determined that PyTorch is the better framework to implement custom models and make changes to existing models. Our team researched current implementations which overlap with our proposed architecture, and sought to integrate the existing codebase with our implementation. 

For implementation of the Variational Graph Autoencoder, we utilized scVI due to its compatibility with PyTorch. For the implementation of the Variational Graph Autoencoder, we followed the implementation of SEDR. Both these repositories were integrated and we established modules and the appropriate class definitions to integrated these modules into our code. We followed a modular, object-oriented approach to ensure organization and scalability of our framework. All related implementation and tests can be found on our team's \href{https://github.com/2255-Spatial-Transcriptomics}{github repository}.  

The pseudo-code for our implementation to train the latent spaces using three steps described in Section~\ref{sec:projectpipeline} is shown in the next page.

\label{pseudocode}
\begin{algorithm}
\label{pseudocode}
\setstretch{1.35}
\caption{The 3 step training procedure}\label{alg:cap}
\begin{algorithmic}
\State \textbf{Prepare:} 
$X_{\textrm{sc, exp}}^{2000}$, \quad 
$X_{\textrm{sc, exp}}^{500}$, \quad \Comment{expression datasets for SC} \\

\hspace{1.8cm} $X_{\textrm{st, exp}}^{500}$, \quad
$X_{\textrm{st, sp}}^{500}$, \quad
\Comment{expression and spatial datasets for ST}
\For {e in s1\_epochs}
\State \textbf{Train:} VAE $f_\textrm{vae1}$
s.t. reconstruction loss of $\widetilde{X}_{\textrm{sc, exp}}^{2000} = f^{-1}_\textrm{vae1}(f_\textrm{vae1}(X_{\textrm{sc, exp}}^{2000}))$ is minimized
\EndFor
\State \textbf{Fix: } $f_\textrm{vae1}(X_{\textrm{sc, exp}}^{2000}) = Z_{\textrm{sc, exp}}^{2000} \rightarrow \bar{Z}_{\textrm{sc, exp}}^{2000} $
\For {\textrm{e in s2\_epochs}}
\State $Z_{\textrm{sc, exp}}^{500} = f_\textrm{vae2}(X_{\textrm{sc, exp}}^{500})$ 
\State $Z_{\textrm{st, exp}}^{500} = f_\textrm{vae3}(X_{\textrm{st, exp}}^{500})$ 
\While{$\textbf{$\mathcal{D}$}(Z_{\textrm{sc, exp}}^{500}, Z_{\textrm{st, exp}}^{500}) < \alpha$ for a max of T iterations}
    \State \textbf{Train: } discriminator \textbf{$\mathcal{D}$} to improve accuracy ($L_2$ of differentiating between latent spaces \\
    \hspace{2.5cm} $L_2 = \textbf{$\mathcal{D}$}(Z_{\textrm{sc, exp}}^{500}, Z_{\textrm{st, exp}}^{500})$
\EndWhile
\For {i in s2b\_epochs}

\State $L_1 = \textrm{dist}(Z_{\textrm{sc, exp}}^{500}, \bar{Z}_{\textrm{sc, exp}}^{2000})$ \Comment{Similarity between SC expression latent spaces}
\State $L_2 = \textbf{$\mathcal{D}$}(Z_{\textrm{sc, exp}}^{500}, Z_{\textrm{st, exp}}^{500})$ \Comment{Similarity between SC and ST latent space}
\State \textbf{Train: } VAE $f_\textrm{vae2}$ by minimizing weighted sum of $L_1$, $L_2$, \\ \hspace{2.5cm} and reconstruction loss of $\widetilde{X}_{\textrm{sc, exp}}^{500} = f^{-1}_\textrm{vae2}(f_\textrm{vae2}(X_{\textrm{sc, exp}}^{500}))$
\State \textbf{Train: } VAE $f_\textrm{vae3}$ by minimizing weighted sum of $L_2$ \\
\hspace{2.5cm} and reconstruction loss of $\widetilde{X}_{\textrm{st, exp}}^{500} = f^{-1}_\textrm{vae3}(f_\textrm{vae3}(X_{\textrm{st, exp}}^{500}))$
\EndFor
\EndFor
\State \textbf{Fix:} $f_\textrm{vae3}(X_{\textrm{st, exp}}^{500}) = Z_{\textrm{st, exp}}^{500} \rightarrow \bar{Z}_{\textrm{st, exp}}^{500} $
\For {e in s3\_epochs}
\State $Z_{\textrm{st, exp+sp}}^{500} = f_\textrm{vgae}(X_{\textrm{st, exp}}^{500}, X_{\textrm{st, sp}}^{500})$
\State $L_3 = \textrm{dist}(Z_{\textrm{st, exp+sp}}^{500}, \bar{Z}_{\textrm{st, exp}}^{500})$ 
\Comment{Similarity to ST expression latent space}
\State \textbf{Train: } VGAE $f_\textrm{vgae}$ by minimizing weighted sum of $L_3$ \\ 
\hspace{2.5cm} and reconstruction loss of $\widetilde{X}_{\textrm{st, exp}}^{500}, \widetilde{X}_{\textrm{st, sp}}^{500} = f_\textrm{vgae}^{-1}(f_\textrm{vgae}(X_{\textrm{st, exp}}^{500}, X_{\textrm{st, sp}}^{500}))$
\EndFor

\end{algorithmic}
\end{algorithm}
% \section{Benchmarking Variational Autoencoder}
% \subsection{Overview}

\label{pseudocode2}
\begin{algorithm}
\label{pseudocode2}
\setstretch{1.35}
\caption{The 3 step training procedure}\label{alg:cap}
\begin{algorithmic}
\State \textbf{Prepare:} 
$X_{\textrm{sc, exp}}^{2000}$, \quad 
$X_{\textrm{sc, exp}}^{500}$, \quad \Comment{expression datasets for SC} \\

\hspace{1.8cm} $X_{\textrm{st, exp}}^{500}$, \quad
$X_{\textrm{st, sp}}^{500}$, \quad
\Comment{expression and spatial datasets for ST}
\For {e in s1\_epochs}
\State \textbf{Train:} VAE $f_\textrm{vae1}$
s.t. reconstruction loss of $\widetilde{X}_{\textrm{sc, exp}}^{2000} = f^{-1}_\textrm{vae1}(f_\textrm{vae1}(X_{\textrm{sc, exp}}^{2000}))$ is minimized
\EndFor
\State \textbf{Fix: } $f_\textrm{vae1}(X_{\textrm{sc, exp}}^{2000}) = Z_{\textrm{sc, exp}}^{2000} \rightarrow \bar{Z}_{\textrm{sc, exp}}^{2000} $
\For {\textrm{e in s2\_epochs}}
\State $Z_{\textrm{sc, exp}}^{500} = f_\textrm{vae2}(X_{\textrm{sc, exp}}^{500})$ 
\State $Z_{\textrm{st, exp}}^{500} = f_\textrm{vae3}(X_{\textrm{st, exp}}^{500})$ 
\While{$\textbf{$\mathcal{D}$}(Z_{\textrm{sc, exp}}^{500}, Z_{\textrm{st, exp}}^{500}) < \alpha$ for a max of T iterations}
    \State \textbf{Train: } discriminator \textbf{$\mathcal{D}$} to improve accuracy ($L_2$ of differentiating between latent spaces \\
    \hspace{2.5cm} $L_2 = \textbf{$\mathcal{D}$}(Z_{\textrm{sc, exp}}^{500}, Z_{\textrm{st, exp}}^{500})$
\EndWhile
\For {i in s2b\_epochs}

\State $L_1 = \textrm{dist}(Z_{\textrm{sc, exp}}^{500}, \bar{Z}_{\textrm{sc, exp}}^{2000})$ \Comment{Similarity between SC expression latent spaces}
\State $L_2 = \textbf{$\mathcal{D}$}(Z_{\textrm{sc, exp}}^{500}, Z_{\textrm{st, exp}}^{500})$ \Comment{Similarity between SC and ST latent space}
\State \textbf{Train: } VAE $f_\textrm{vae2}$ by minimizing weighted sum of $L_1$, $L_2$, \\ \hspace{2.5cm} and reconstruction loss of $\widetilde{X}_{\textrm{sc, exp}}^{500} = f^{-1}_\textrm{vae2}(f_\textrm{vae2}(X_{\textrm{sc, exp}}^{500}))$
\State \textbf{Train: } VAE $f_\textrm{vae3}$ by minimizing weighted sum of $L_2$ \\
\hspace{2.5cm} and reconstruction loss of $\widetilde{X}_{\textrm{st, exp}}^{500} = f^{-1}_\textrm{vae3}(f_\textrm{vae3}(X_{\textrm{st, exp}}^{500}))$
\EndFor
\EndFor
\State \textbf{Fix:} $f_\textrm{vae3}(X_{\textrm{st, exp}}^{500}) = Z_{\textrm{st, exp}}^{500} \rightarrow \bar{Z}_{\textrm{st, exp}}^{500} $
\For {e in s3\_epochs}
\State $Z_{\textrm{st, exp+sp}}^{500} = f_\textrm{vgae}(X_{\textrm{st, exp}}^{500}, X_{\textrm{st, sp}}^{500})$
\State $L_3 = \textrm{dist}(Z_{\textrm{st, exp+sp}}^{500}, \bar{Z}_{\textrm{st, exp}}^{500})$ 
\Comment{Similarity to ST expression latent space}
\State \textbf{Train: } VGAE $f_\textrm{vgae}$ by minimizing weighted sum of $L_3$ \\ 
\hspace{2.5cm} and reconstruction loss of $\widetilde{X}_{\textrm{st, exp}}^{500}, \widetilde{X}_{\textrm{st, sp}}^{500} = f_\textrm{vgae}^{-1}(f_\textrm{vgae}(X_{\textrm{st, exp}}^{500}, X_{\textrm{st, sp}}^{500}))$
\EndFor

\end{algorithmic}
\end{algorithm}

\chpt{Results} \label{chpt:results}
\section{Benchmarking Results}

We present the benchmarking results for each method in the following tables. For scPhere, we observed the latent representations for 2, 10 and 20 dimensions on a hyper-sphere. We assumed a negative-binomial distribution for the observation distribution, and we computed the latent space distribution, which is assumed to be a von Mises-Fisher distribution. \text{Avg CV Acc} stores the average 4-fold cross-validation accuracy.

\begin{table}[h]
\resizebox{\columnwidth}{!}{%
\begin{tabular}{|l|llllll|}
\hline
\multicolumn{1}{|c|}{\multirow{3}{*}{Datasets}} & \multicolumn{6}{c|}{scPhere Accuracy} \\ \cline{2-7} 
\multicolumn{1}{|c|}{} & \multicolumn{2}{c|}{2D} & \multicolumn{2}{c|}{10D} & \multicolumn{2}{c|}{20D} \\ \cline{2-7} 
\multicolumn{1}{|c|}{} & \multicolumn{1}{l|}{KNN Acc} & \multicolumn{1}{l|}{Avg CV Acc} & \multicolumn{1}{l|}{KNN Acc} & \multicolumn{1}{l|}{Avg CV Acc} & \multicolumn{1}{l|}{KNN Acc} & Avg CV Acc \\ \hline
Dataset   A & \multicolumn{1}{l|}{88.90\%} & \multicolumn{1}{l|}{89.86\%} & \multicolumn{1}{l|}{89.87\%} & \multicolumn{1}{l|}{89.87\%} & \multicolumn{1}{l|}{71.18\%} & 76.61\% \\ \hline
Dataset   B & \multicolumn{1}{l|}{95.94\%} & \multicolumn{1}{l|}{95.07\%} & \multicolumn{1}{l|}{94.78\%} & \multicolumn{1}{l|}{95.56\%} & \multicolumn{1}{l|}{95.36\%} & 95.79\% \\ \hline
Dataset   C & \multicolumn{1}{l|}{81.81\%} & \multicolumn{1}{l|}{76.33\%} & \multicolumn{1}{l|}{86.28\%} & \multicolumn{1}{l|}{75.77\%} & \multicolumn{1}{l|}{86.46\%} & 76.20\% \\ \hline
\end{tabular}%
}
\caption{scPhere kNN accuracy using 4-fold cross-validation.}
\end{table}

For scDHA, the latent dimension $m$ is predetermined in the library to be $25$ if the dataset has more than $50000$ genes and $15$ otherwise since varying $m$ between $10$ and $20$ does not alter the analysis results \cite[p.8]{Tran799817}. All our datasets have less than $50000$ genes, thus the following analysis has a latent dimension of $15$. For latent space distribution plots, kNN accuracy v.s. $k$ plots, and confusion matrices, please see \nameref{scDHA_sup}.

\begin{table}[h]
\centering
\begin{tabular}{|l|ll|}
\hline
\multicolumn{1}{|c|}{} & \multicolumn{2}{c|}{scDHA Accuracy} \\ \cline{2-3} 
\multicolumn{1}{|c|}{} & \multicolumn{2}{c|}{15D} \\ \cline{2-3} 
\multicolumn{1}{|c|}{\multirow{-3}{*}{Datasets}} & \multicolumn{1}{l|}{Avg CV Acc} & ARI \\ \hline
Dataset A & \multicolumn{1}{l|}{{\color[HTML]{000000} 84.17\%}} & 0.54 \\ \hline
Dataset B & \multicolumn{1}{l|}{{\color[HTML]{000000} 94.2\%}} & 0.32 \\ \hline
Dataset C & \multicolumn{1}{l|}{{\color[HTML]{000000} 86.7\%}} & 0.17 \\ \hline
\end{tabular}
\caption{scDHA kNN 4-fold cross-validation accuracy with ARI score}
\end{table}

For VASC, we observed the normal latent representations for 2, 10 and 20 dimensions. We assumed a multi-dimensional normal prior distribution for the observation distribution, and we computed the posterior distribution.

\begin{table}[h]
\resizebox{\columnwidth}{!}{%
\begin{tabular}{|l|llllll|}
\hline
\multicolumn{1}{|c|}{\multirow{3}{*}{Datasets}} & \multicolumn{6}{c|}{VASC Accuracy} \\ \cline{2-7} 
\multicolumn{1}{|c|}{} & \multicolumn{2}{c|}{2D} & \multicolumn{2}{c|}{10D} & \multicolumn{2}{c|}{20D} \\ \cline{2-7} 
\multicolumn{1}{|c|}{} & \multicolumn{1}{l|}{KNN Acc} & \multicolumn{1}{l|}{Avg CV Acc} & \multicolumn{1}{l|}{KNN Acc} & \multicolumn{1}{l|}{Avg CV Acc} & \multicolumn{1}{l|}{KNN Acc} & Avg CV Acc \\ \hline
Dataset   A & \multicolumn{1}{l|}{64.41\%} & \multicolumn{1}{l|}{70.57\%} & \multicolumn{1}{l|}{72.74\%} & \multicolumn{1}{l|}{77.37\%} & \multicolumn{1}{l|}{69.96\%} & 75.41\% \\ \hline
Dataset   B & \multicolumn{1}{l|}{84.06\%} & \multicolumn{1}{l|}{87.01\%} & \multicolumn{1}{l|}{86.09\%} & \multicolumn{1}{l|}{87.88\%} & \multicolumn{1}{l|}{86.67\%} & 87.74\% \\ \hline
Dataset   C & \multicolumn{1}{l|}{30.64\%} & \multicolumn{1}{l|}{37.51\%} & \multicolumn{1}{l|}{54.24\%} & \multicolumn{1}{l|}{55.61\%} & \multicolumn{1}{l|}{54.98\%} & 56.82\% \\ \hline
\end{tabular}%
}
\caption{VASC kNN accuracy using 4-fold cross-validation.}
\end{table}

For scVI, we assume a normal distribution for the latent space representation. We observe the latent space representation for 2, 10 and 20D on a normal distribution. We see that scVI performs similar to scPhere for 2D and 10D representations for Datasets A and B. However, scVI's performance seems to decrease as we increase the number of latent dimensions. This is not problematic for us since we want a low dimensional representation of our data and 10D representation seems ideal for this case. In addition to this, scVI is implemented in PyTorch and thus is compatible with our VGAE models.  
\begin{table}[h]
\resizebox{\columnwidth}{!}{%
\begin{tabular}{|l|llllll|}
\hline
\multicolumn{1}{|c|}{\multirow{3}{*}{Datasets}} & \multicolumn{6}{c|}{scVI Accuracy} \\ \cline{2-7} 
\multicolumn{1}{|c|}{} & \multicolumn{2}{c|}{2D} & \multicolumn{2}{c|}{10D} & \multicolumn{2}{c|}{20D} \\ \cline{2-7} 
\multicolumn{1}{|c|}{} & \multicolumn{1}{l|}{KNN Acc} & \multicolumn{1}{l|}{Avg CV Acc} & \multicolumn{1}{l|}{KNN Acc} & \multicolumn{1}{l|}{Avg CV Acc} & \multicolumn{1}{l|}{KNN Acc} & Avg CV Acc \\ \hline
Dataset   A & \multicolumn{1}{l|}{89.51\%} & \multicolumn{1}{l|}{89.86\%} & \multicolumn{1}{l|}{83.96\%} & \multicolumn{1}{l|}{86.12\%} & \multicolumn{1}{l|}{25.934\%} & 32.95\% \\ \hline
Dataset   B & \multicolumn{1}{l|}{91.88\%} & \multicolumn{1}{l|}{94.19\%} & \multicolumn{1}{l|}{93.33\%} & \multicolumn{1}{l|}{93.105\%} & \multicolumn{1}{l|}{44.637\%} & 53.628\% \\ \hline
Dataset   C & \multicolumn{1}{l|}{68.74\%} & \multicolumn{1}{l|}{67.17\%} & \multicolumn{1}{l|}{85.77\%} & \multicolumn{1}{l|}{75.256\%} & \multicolumn{1}{l|}{70.20\%} & 54.31\% \\ \hline
\end{tabular}%
}
\caption{scVI kNN accuracy using 4-fold cross-validation.}
\end{table}

DRA uses the most novel approach, generative adversarial networks, out of the four methods. Similar to other methods, we can obtain the latent matrix $z$ for all datasets. However, the results for the latent representations were not ideal. We obtained an average 4-fold cross-validation accuracy of $38\%$ for Dataset A and sub-optimal performances for Datasets B and C. Therefore, we did not move forward with this method. 

\subsection{VAE Model Selection} \label{sec:vae-model-selection}
The results of our top three performing models are shown in Figure \ref{fig:VAEComparison}. 

\begin{figure}[H]
    \centering
    \includegraphics[scale=0.45]{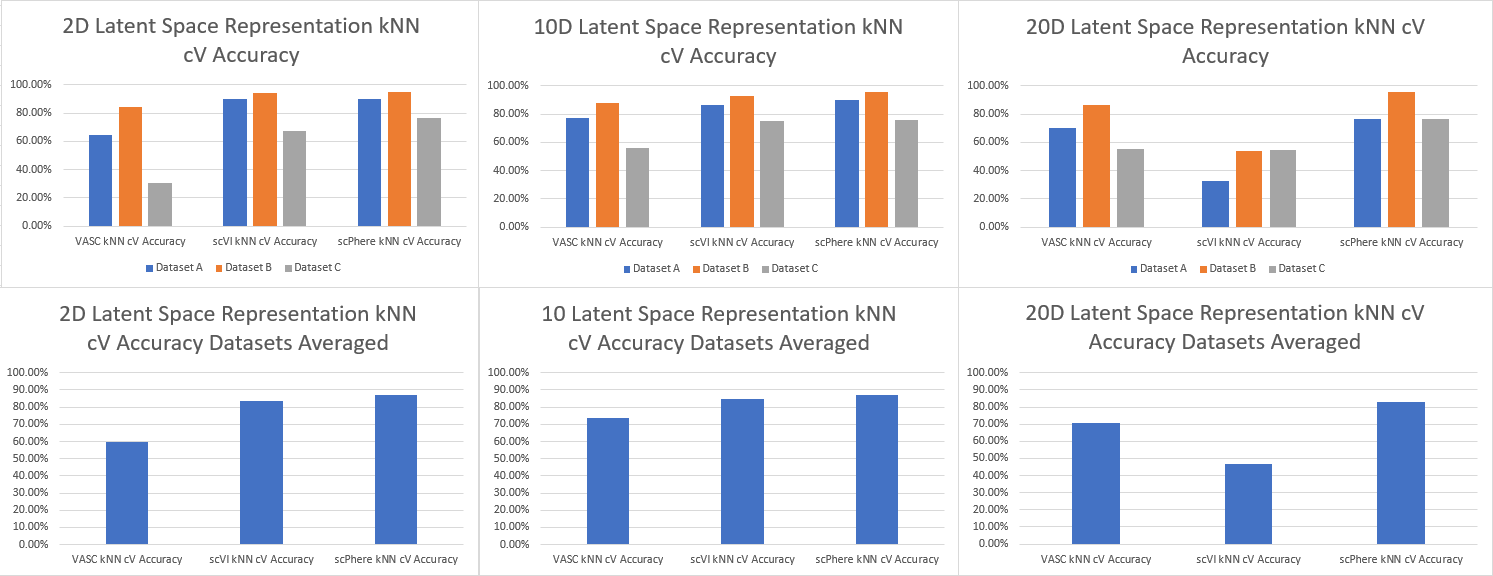}
   \caption{Top three VAE Comparisons} 
   \label{fig:VAEComparison}
\end{figure}  

We see that overall, scVI tends to perform better compared to VASC for 2D and 10D latent representation but performs significantly worse for 20D latent representation. scPhere performs the best on average for all the dimensions on average. 
We picked \textbf{scVI} for our VAE since it performs about the same as scPhere for 2D and 10D latent space representation and in addition to this, scVI is implemented in PyTorch so it is easier to integrate with our VGAE model which is also implemented in PyTorch.

\section{Discriminator Implementation} \label{sec:disc_imp}
For the discriminator, we implement a standard binary classifier in PyTorch from scratch. We implemented a model with 3 hidden layers, each with 128 nodes, RELU activation, Binary Cross Entropy with Logits as the loss function, and the Adam optimizer. The discriminator network was implemented as its own class in our code repository to allow easy modifications to model hyper-parameters for future fine-tuning. To verify the performance of our discriminator,  on two latent spaces generated from scVI, this time using 10 dimensional latent space features, as shown in figure \ref{}. We visualize the results by plotting the 2 principal components of both datasets, and note that the discriminator learns to distinguish the high dimensional data with great precision. Here the discriminator was trained also for 30 epochs. 
\begin{figure}[H]
    \centering
    \includegraphics[scale=0.35]{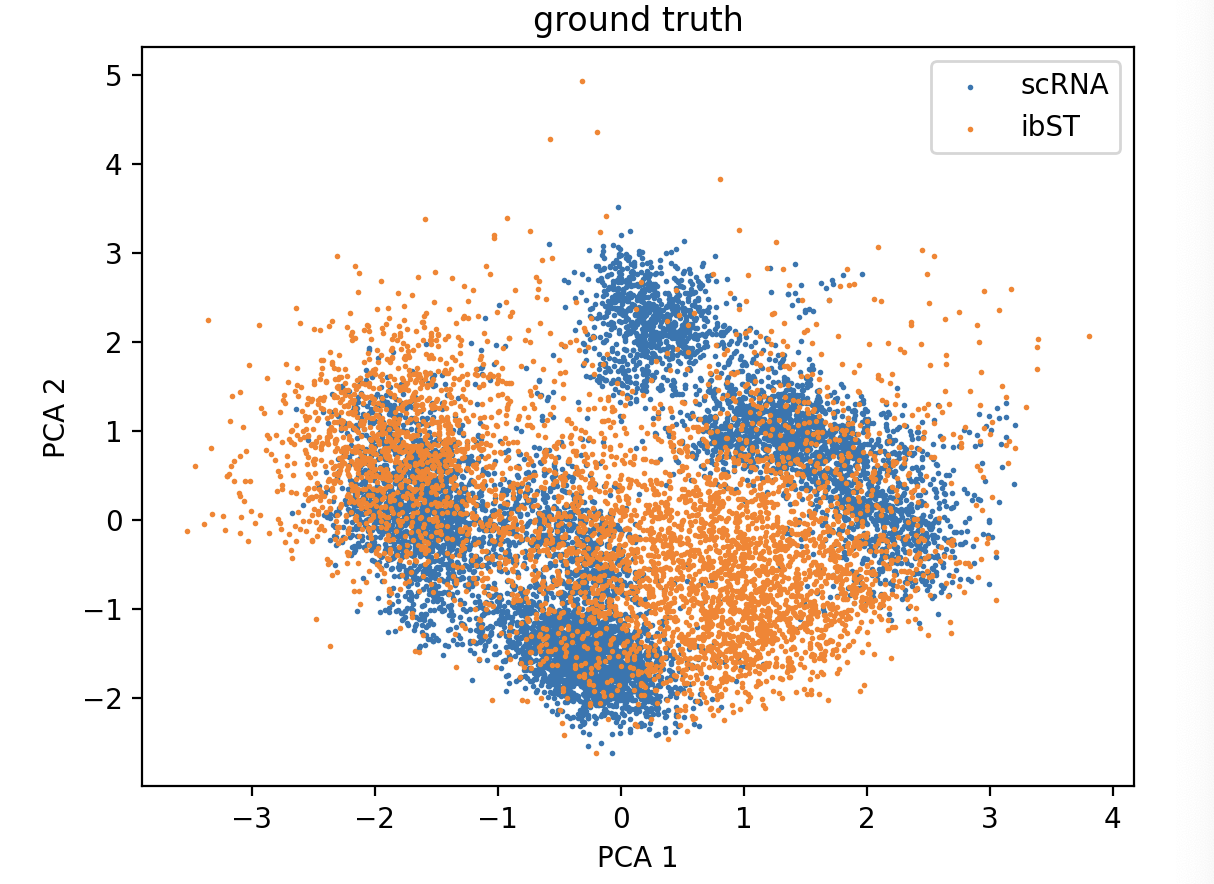}
    \includegraphics[scale=0.35]{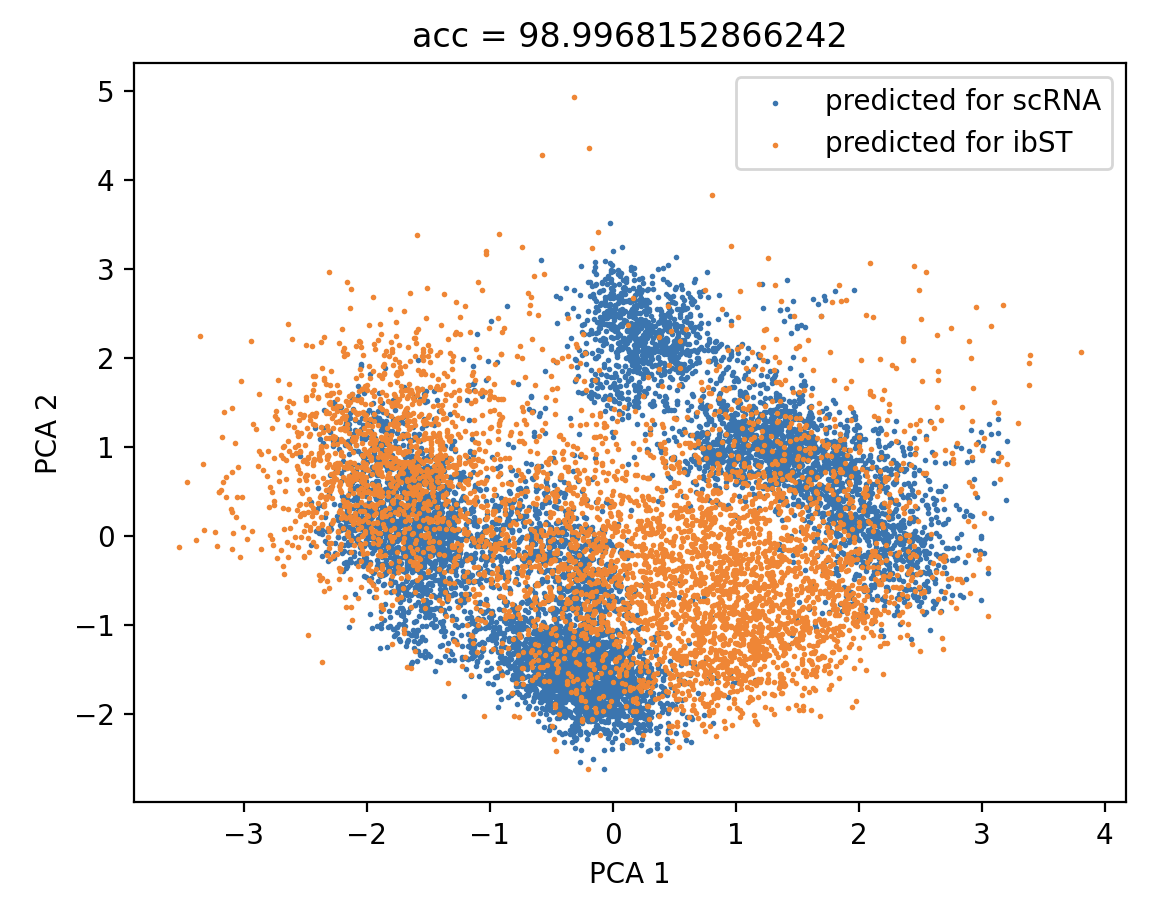}
    \caption{Results of discriminator prediction results after training for 10 epochs. The dataset consists of the latent representations of a single cell RNA sequencing dataset (collected from Allen Merfish) and a spatial transcriptomics dataset (collected from 10x Genomics) for a human brain tissue. We note that the discriminator is able to successfully distinguish between the two latent spaces with high accuracy. When projecting the first two principal components of both datasets, we see that there is significant overlap between the latent distributions.}
    \label{fig:discriminator-latent}
\end{figure}
\section{Results for SEDR} \label{chpt:results2}

The SEDR framework performs latent mapping of spatial transcriptomics datasets and produces a latent representation of both the spatial graph and gene expression. We make modifications on top of this architecture and introduce an additional hidden layer within the network, which combines the latent spaces of the spatial information and gene expression information into a single latent space. We denote the original SEDR as SEDR\_v1 and our modified model as SEDR\_v2.

\begin{figure}[H]
    \centering
    \includegraphics[scale=0.6]{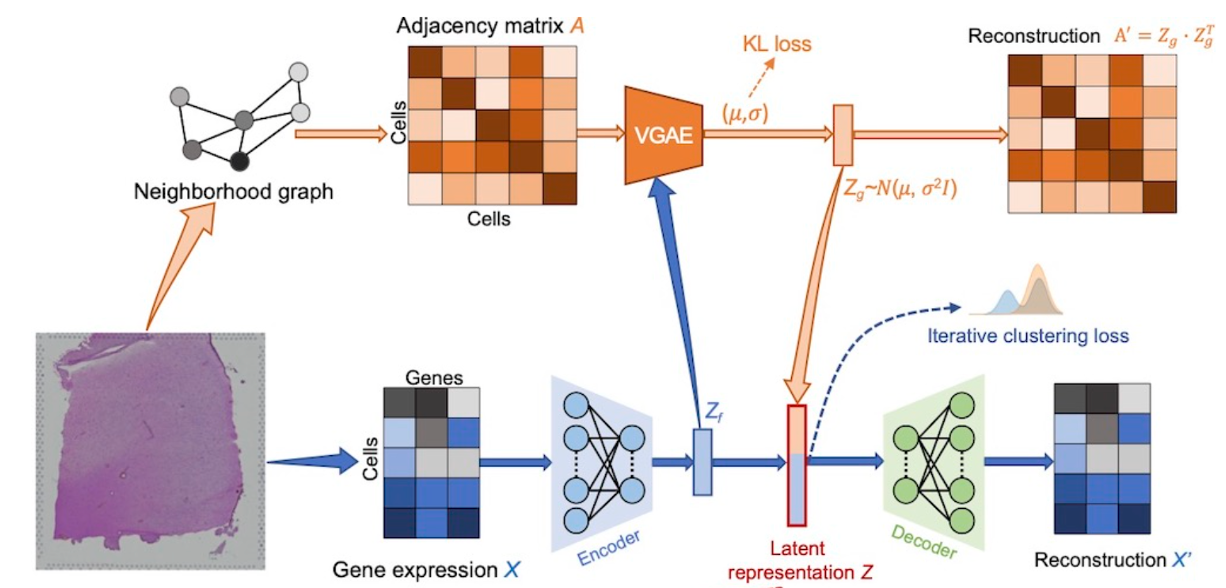}
    \includegraphics[scale=0.6]{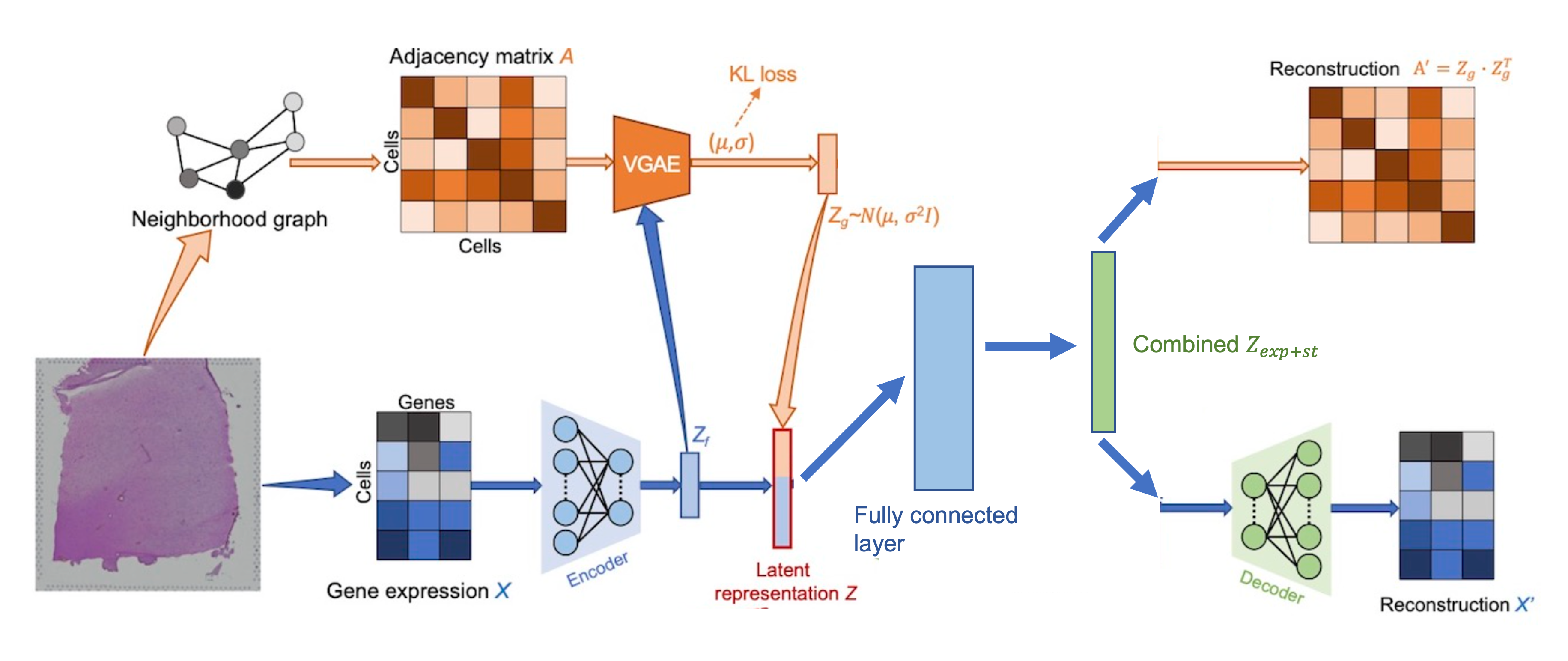}
    \caption{Comparing the original SEDR framework (top) \citep{sedr} to our implementation (bottom). After concatenation of the latent spaces of spatial and expression data, we then merge the two into a single latent space through the use of a fully connected layer. }
\end{figure}

We compare the results of SEDR\_v2 with the original model and conclude that the combined latent representation via the fully connected layer is feasible for retaining the learned information. The results can be seen in figure \ref{sedr_compare}. Further tests are performed which confirm the feasibility of this model and can be seen on our github repo. 
\begin{figure}[H]
    \centering
    \includegraphics[scale=0.6]{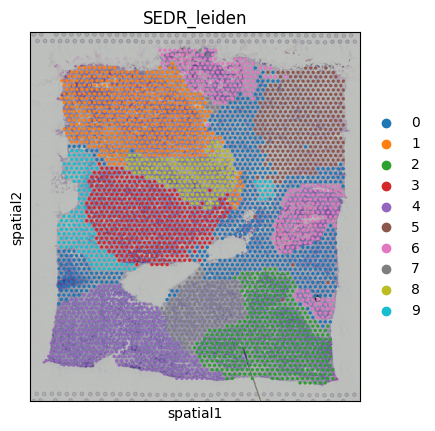}
    \includegraphics[scale=0.6]{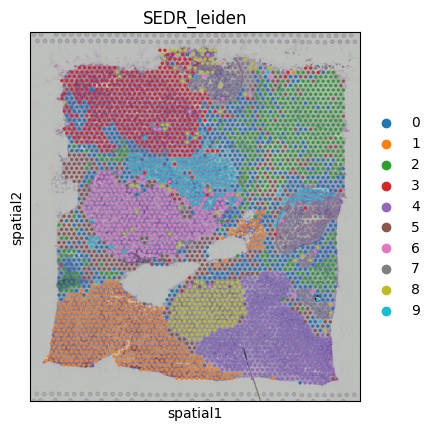}
    \caption{Comparing the results of SEDR\_v1 (left) versus SEDR\_v2 (right) results for a breast cancer dataset. Both models were trained for 500 epochs with DEC loss. Results obtained from original model achieved a loss of 212.9 while our modified model achieves a loss of 222.6}
    \label{sedr_compare}
\end{figure}

Next we compare the results generated using our modified SEDR architecture with the original tissue sample. Our test for a human brain tissue sample is shown in figure \ref{brain_tissue_compare}. 
\begin{figure}[H]
    \centering
    \includegraphics[scale=0.6]{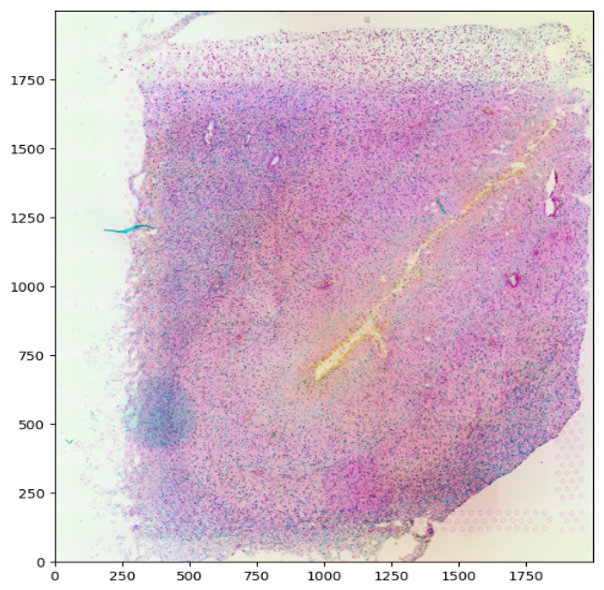}
    \includegraphics[scale=0.6]{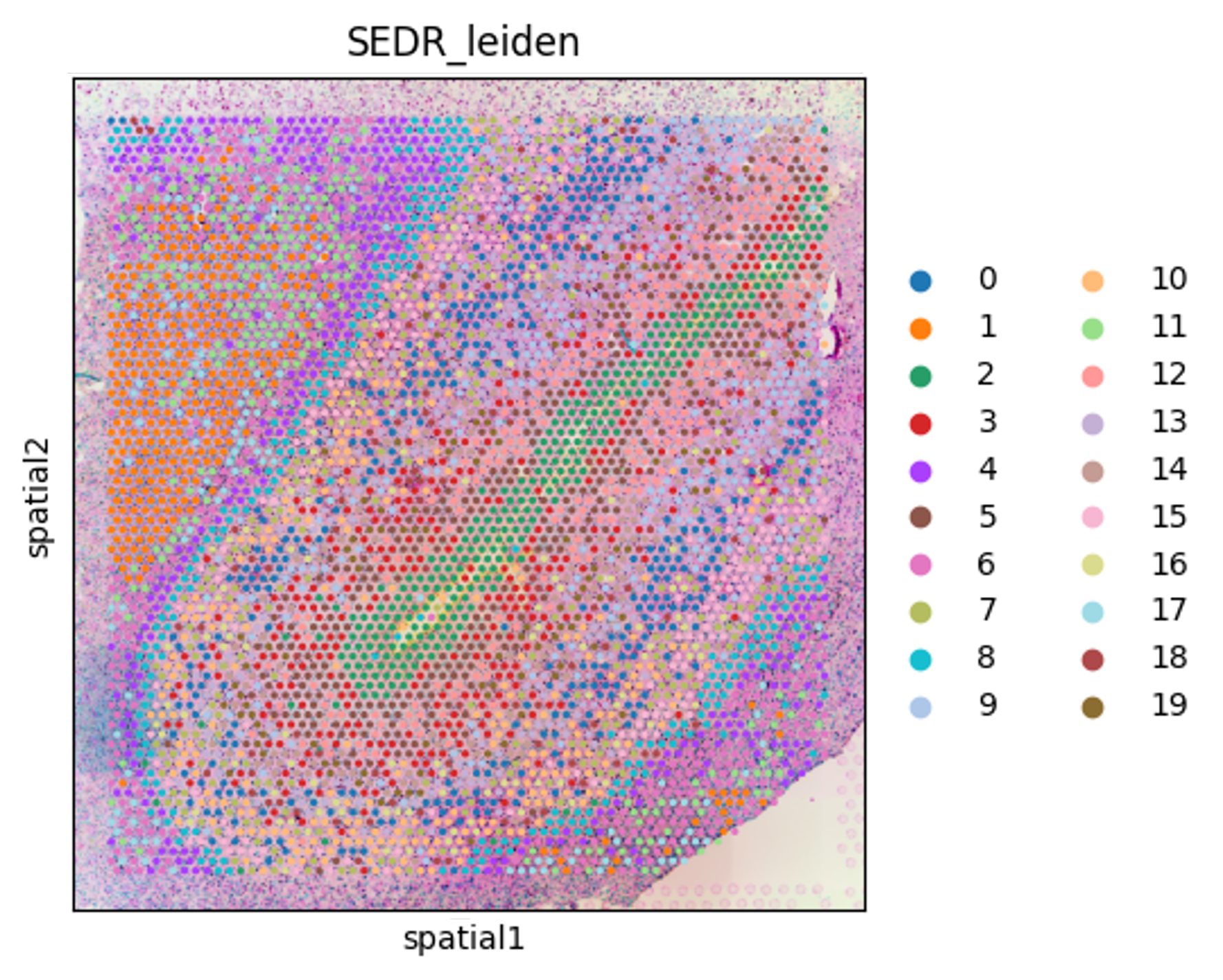}
    \caption{Comparison between the mapping of cells from the latent space of our model (right) as compared to the original tissue (left). We see that the combined latent space of spatial and expression data provides a good representation of the ground truth cell groupings in the tissue. }
    \label{brain_tissue_compare}
\end{figure}
\chapter{Conclusions}\label{chpt:Conclusion}
 Overall, we explored various Variational Autoencoders in this project to obtain a good latent space representation of scRNA data. We explored the architecture of these different VAEs and compared their performances on similar datasets. Based on the results and also for easier implementation, we decided to pick the VAE, scVI, to find a latent space representation for the scRNA data (Section~\ref{sec:vae-model-selection}). For our spatial transcriptomics dataset, we used a well-known VGAE, SEDR, to find the latent space representation of the spatial transcriptomics dataset. We modified the existing SEDR architecture to introduce additional hidden layers in SEDR's encoder network to generate a shared latent space for the gene expression and spatial information of spatial transcriptomics data (Section~\ref{chpt:results2}). We obtained good latent space representations for scRNA and spatial transcriptomics datasets using these architectures. In addition, we were able to develop a discriminator model and utilize it to minimize the adversarial loss between the latent space of the spatial transcriptomics gene expression data and the scRNA data. (Section~\ref{sec:disc_imp})

 Although we could implement the two pipelines of our final model, the latent space of the scRNA dataset and the latent space of the Spatial Transcriptomics dataset, due to time constraints, we needed more time to finish mapping the two latent spaces. Once the mapping of the latent spaces is established, we can proceed to implement the final stage of our project, which is to infer the spatial information of the cells based on their gene expression information.

\chapter{Recommendations}\label{chpt:Recommendations}
We believe that this project has the potential to branch out in multiple directions and lead to exciting discoveries. We recommend future researchers to explore the following implementations and ideas:

\section{Perform more Quantitative and Qualitative Evaluations}
Although we investigated in common VAE/VGAE evaluation metrics (kNN, ARI) and machine learning performance assessments from literature, there are many more robust analysis that can be done on the overall performance and accuracy of the pipeline. 

\section{Hyperbolic and Hyperspherical Latent Prior and Batch Correction}
According to our benchmarking results, scPhere \citep{scphere} has the best performance among all VAE methods due to its enforcement of the prior distribution and batch correction considerations. Due to implementation constraints (see \nameref{sec:vae-model-selection}), we selected scVI over scPhere because the former was implemented in PyTorch, which is consistent with SEDR. Enforcing a hyperbolic or hyperspherical latent prior distribution will reduce distortion and cluttering of the latent representation \citep{scphere}. Additionally, batch correction considers different biological or external factors when training the VAE. Exploring how the whole pipeline would perform if these considerations were included would be a valuable investigation. 

 \section{Pipeline Validation}
Currently, we only validated our pipeline using a small MERFISH dataset we obtained from Allen Brain Institute. It is necessary that future researchers validate this pipeline using larger datasets and experiment with data from different tissues, different number of cells/genes and different machine learning parameters (epoch, layers, activation functions, etc.). 

\chapter{Deliverables}\label{chpt:Deliverables}
We agreed to deliver the following deliverables to our project sponsor:
\begin{enumerate}
\item Proof of concept "latent mapping" architecture:
\begin{enumerate}
    \item Fully functional probabilistic machine learning pipeline 
utilizing VAE and VGAE
    \item Encoding of scRNA sequencing data and spatial transcriptomics data into a learned latent space 
    \item Ability to perform latent mapping between latent spaces of our data
    \item Ability to infer spatial information from latent space 
\end{enumerate}
\item Published code-base and documentation of our research
\item Project proposal, presentations, and final report
\end{enumerate}

\end{mainf}
%-------------------------------------------

%%%%% OPTIONAL APPENDICES %%%%%
%-------------------------------------------
% If a single appendix is not applicable, comment out lines 243-249 to hide the optional single appendix.
\begin{appendices}
\multappendices
\chpt{Benchmarking VAE}
\section{Benchmarking Methods}

\subsection{scPhere: Deep Generative Model Embedding on Hyperspheres and Hyperbolic Spaces} \label{scphere_overview}
Current dimension reduction techniques tend to map high-dimensional data into a low-dimensional Cartesian latent space, which results in a cluttered, inaccurate latent representation of the data. \textbf{scPhere}, a deep generative model designed by \citeauthor{scphere}, is dedicated to mitigating this issue by mapping the RNA sequencing data into a hyperspherical or hyperbolic latent space. The resulting latent representation would be more naturally distributed and more accurately represent the high dimensional data. scPhere also implements multi-level batch correction. \cite{scphere} Single-cell profiles in datasets are usually impacted by diverse factors including technical batch effects in different experiments with different lab protocols and also biological factors which include inter-individual variations, sex, diseases and tissue location. scPhere can learn models of data with multiple variables. In addition to this, scPhere preserves the structure of sc-RNA data in low dimensional spaces even. With all of these advantages, scPhere can help to find an optimal latent space representation for single-cell RNA sequencing data.

\subsection{scDHA: Single-Cell Decomposition using Hierarchical Autoencoder} \label{scDHA_overview}
\textbf{scDHA}, single-cell decomposition using hierarchical autoencoder designed by \citeauthor{Tran799817}, uses a two-stage autoencoder to reduce noise and dimensions. The first stage uses a non-negative kernel autoencoder to obtain the part-based representation of the initial data by removing data that has a small contribution to the latent features. The second stage uses a stacked Bayesian self-learning network to project the data into multiple low-dimensional latent spaces and obtain their latent embeddings. This research implemented an R library and mainly focused on the performance of cell segregation through unsupervised learning, transcriptome landscape visualization, cell classification, and pseudo-time inference. \cite{Tran799817}

\subsection{VASC: Deep Variational Autoencoder} \label{VASC_overview}
\textbf{VASC}, dimensionality reduction with deep Variational autoencoder is a multi-layer generative model for dimensionality reduction. In general, VASC models the distribution of high-dimensional original data $P(X)$ by a set of latent variables $z$. The primary goal of VASC is to find the optimal $z$ by capturing the intrinsic information of the input data. VASC tries to determine the posterior distribution $P(z|X)$ by designing another common distribution family $Q(z|X)$, the variational distribution, to approximate $P(z|X)$. $P(z|X)$ is then approximated from $Q(z|X)$ using the Kullback-Leibler (KL) divergence between the two distributions. Deep neural networks are used in VASC to model the variational distribution $Q(z|X)$. The structure of VASC can be summarized as follows:
\begin{itemize}
    \item \textbf{Input Layer}: This layer takes in the expression matrix from sc-RNA data and applies preprocessing such as log-transform for robust results. It also rescales the expression of every gene by normalizing it with the maximum expression value of an individual gene from the same cell.
    \item \textbf{Dropout Layer}: This layer sets some features to zero during the encoding phase. This introduces random noise in the data.
    \item \textbf{Encoder Network}: This is a three-layer fully connected neural network which carries out dimensionality reduction, reducing the dimension by 4 in each layer. 
    \item \textbf{Latent sampling layer}: Latent variables $z$ were modelled as a normal distribution with standard normal prior $N(0, I)$. This was used to estimate the posterior parameters, $\mu, \ \Sigma$.
    \item \textbf{Decoder network}: The decoder network used the generated latent variables $z$ to recover the original expression matrix, which was designed as a three-layer neural network fully connected with hidden unit dimensions of 32, 128 and 512 and an output layer. 
    \item \textbf{Zero Inflated layer}: The zero-inflated (ZI) layer, which models the dropout events by the distribution $e^{-\widetilde{y}^2}$, where $\widetilde{y}$ is the recovered expression value.

\end{itemize}

\subsection{DRA: Deep Adversarial Variational Autoencoder} \label{DRA_overview}
\textbf{DRA}, dimensionality reduction with adversarial variational autoencoder, is a GAN-based architecture that combines an adversarial autoencoder and a variational autoencoder to reduce dimensions for single-cell RNA sequencing data. \citeauthor{wang_gu_2018} aims to use a novel Dual Matching (AVAE-DM) system, which consists of a generator and two discriminators, to obtain the optimal latent representation of the data. \cite{wang_gu_2018}

\subsection{scVI: Deep generative modeling for single-cell transcriptomics}
\label{scVI_overview}
scVI is a popular software package that uses a variational autoencoder (VAE) for the analysis of single-cell RNA sequencing (scRNA-seq) data. The scVI model learns a low-dimensional representation of the gene expression data that captures the underlying biological variability. The model incorporates a number of design features that make it particularly well-suited to the analysis of scRNA-seq data, including the ability to handle sparse data, batch effects, and missing data. scVI has been shown to outperform existing methods for clustering, visualization, and imputation of scRNA-seq data, making it a valuable tool for researchers in the field. In addition, scVI is open source and available for use by the broader scientific community.

\section{Benchmarking Results}
\subsection{Dataset Cell Type Distribution} \label{dataset_dist}
\begin{figure}[!hp]
\minipage{0.32\textwidth}
  \includegraphics[width=\linewidth]{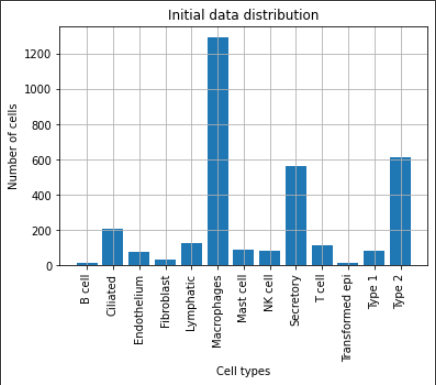}
  \caption{Dataset A}
\endminipage\hfill
\minipage{0.32\textwidth}
  \includegraphics[width=\linewidth]{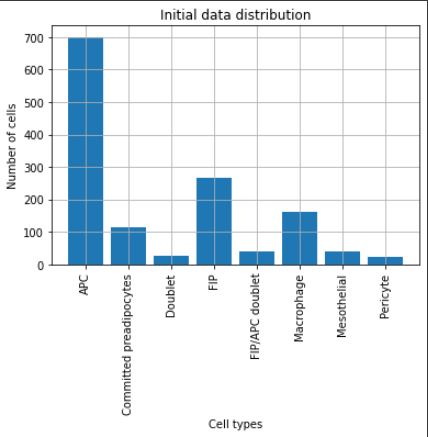}
  \caption{Dataset B}
\endminipage\hfill
\minipage{0.32\textwidth}
  \includegraphics[width=\linewidth]{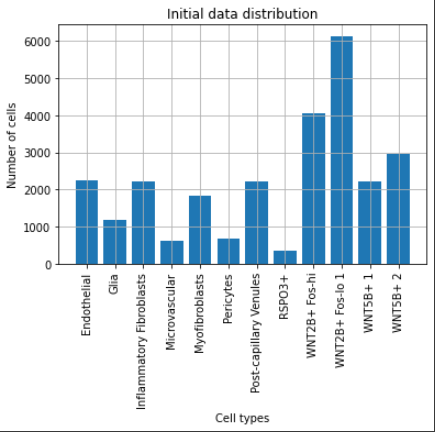}
  \caption{Dataset C}
\endminipage
\end{figure}

\subsection{scPhere Supplemental Figures} \label{scPhere_sup}
     \begin{table}[htb]
     \begin{center}
    \caption{scPhere Results for 2D, 10D, 20D for Dataset A}
     \begin{tabular}{ | c | p{5cm} | p{5cm} |  p{5cm} | }
     \hline
      Dim & kNN Accuracy vs k & 4-fold cross-validation Accuracy & Confusion Matrix \\ \hline
      2-D
      &
      
     \includegraphics[width=0.3\textwidth, height=40mm]{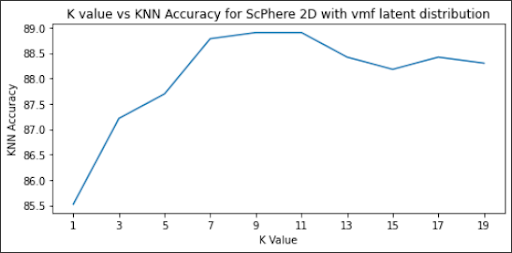}
      & 
    \includegraphics[width=0.3\textwidth, height=40mm]{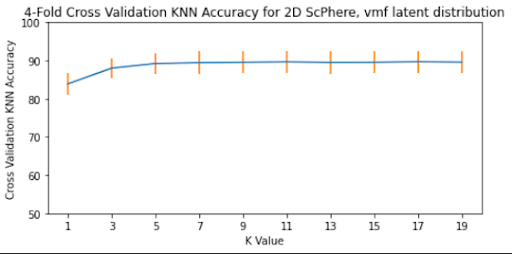}
      & 
    \includegraphics[width=0.3\textwidth, height=40mm]{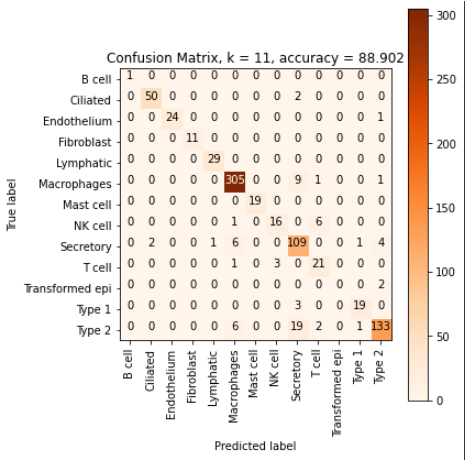}  
      \\ \hline
    10-D
      &
      
     \includegraphics[width=0.3\textwidth, height=40mm]{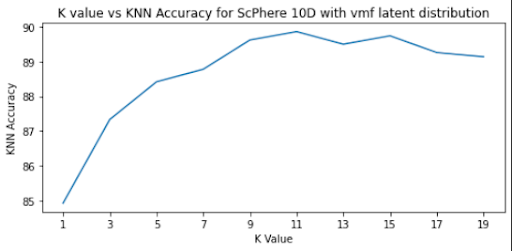}
      & 
    \includegraphics[width=0.3\textwidth, height=40mm]{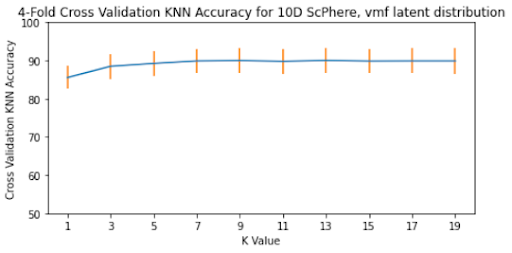}
      & 
    \includegraphics[width=0.3\textwidth, height=40mm]{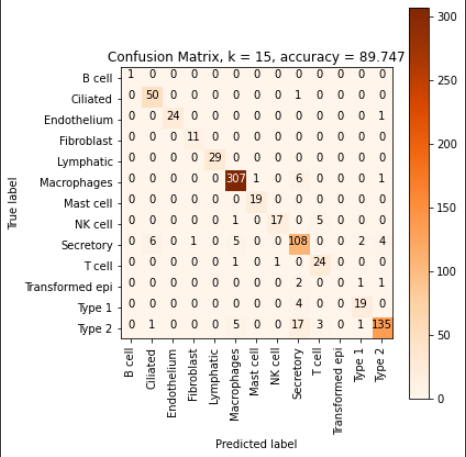}  
    \\ \hline
    20-D
      &
      
     \includegraphics[width=0.3\textwidth, height=40mm]{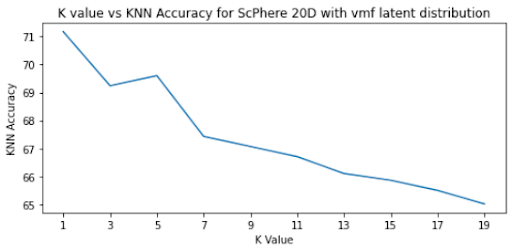}
      & 
    \includegraphics[width=0.3\textwidth, height=40mm]{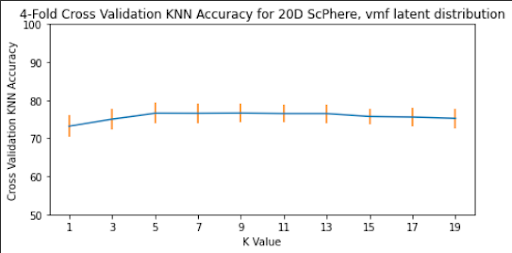}
      & 
    \includegraphics[width=0.3\textwidth, height=40mm]{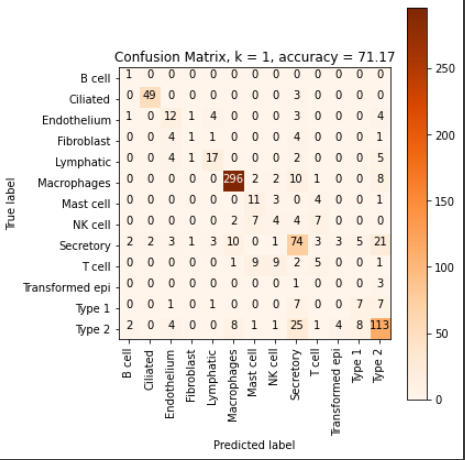}  
    \\ \hline
    
      \end{tabular}
      \label{tbl:myLboro}
      \end{center}
      \end{table}

\newpage
\begin{table}[htb]
     \begin{center}
    \caption{scPhere Results for 2D, 10D, 20D for Dataset B}
     \begin{tabular}{ | c | p{5cm} | p{5cm} |  p{5cm} | }
     \hline
      Dim & kNN Accuracy vs k & 4-fold cross-validation Accuracy & Confusion Matrix \\ \hline
      2-D
      &
      
     \includegraphics[width=0.3\textwidth, height=40mm]{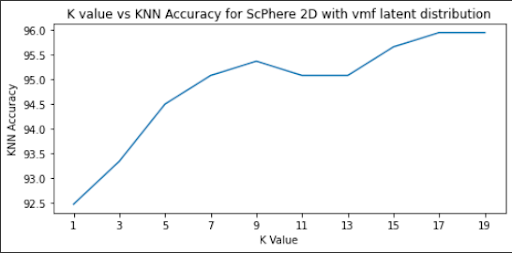}
      & 
    \includegraphics[width=0.3\textwidth, height=40mm]{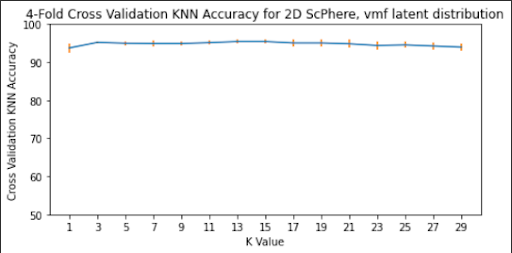}
      & 
    \includegraphics[width=0.3\textwidth, height=40mm]{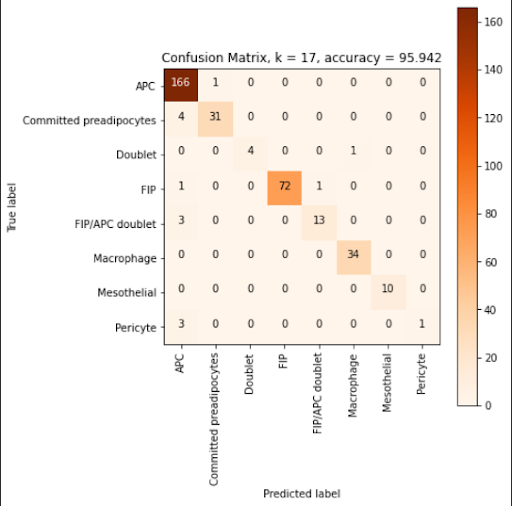}  
      \\ \hline
    10-D
      &
      
     \includegraphics[width=0.3\textwidth, height=40mm]{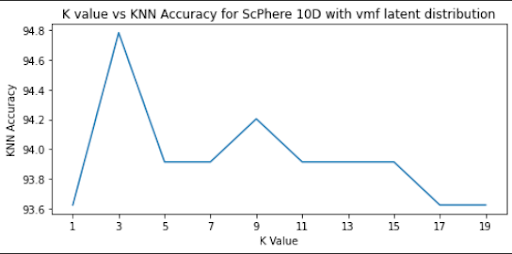}
      & 
    \includegraphics[width=0.3\textwidth, height=40mm]{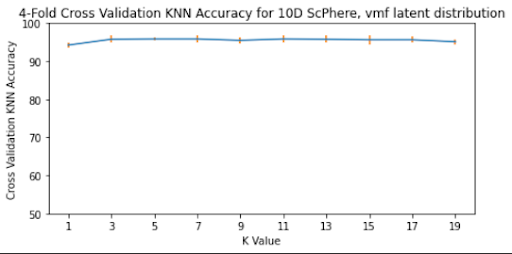}
      & 
    \includegraphics[width=0.3\textwidth, height=40mm]{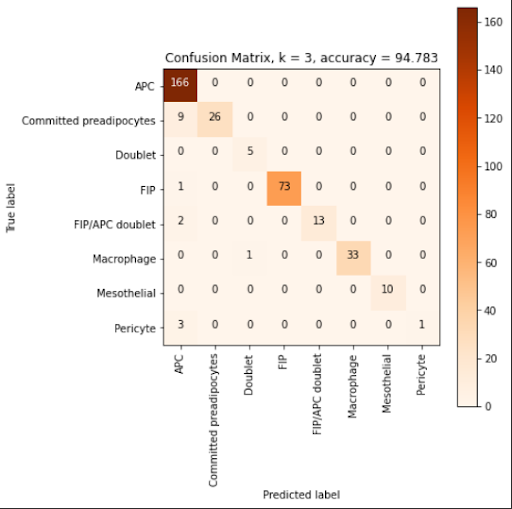}  
    \\ \hline
    20-D
      &
      
     \includegraphics[width=0.3\textwidth, height=40mm]{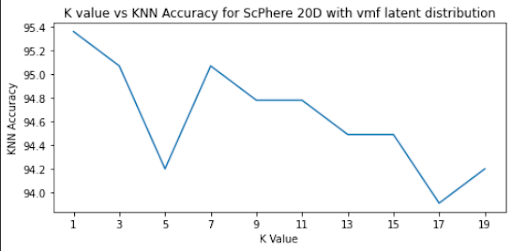}
      & 
    \includegraphics[width=0.3\textwidth, height=40mm]{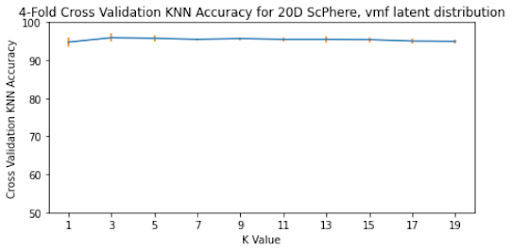}
      & 
    \includegraphics[width=0.3\textwidth, height=40mm]{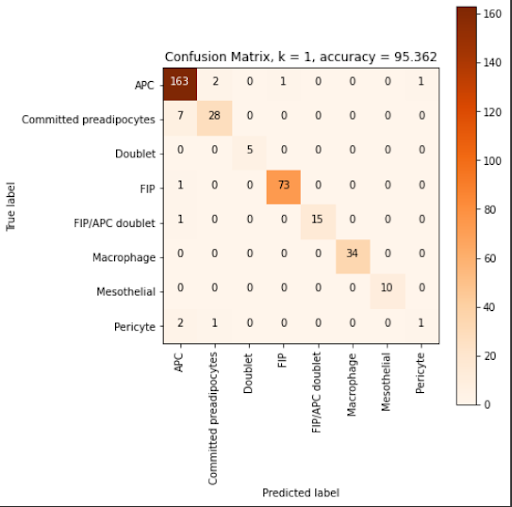}  
    \\ \hline
    
      \end{tabular}
      \label{tbl:myLboro}
      \end{center}
      \end{table}

\newpage
\begin{table}[htb]
     \begin{center}
    \caption{scPhere Results for 2D, 10D, 20D for Dataset C}
     \begin{tabular}{ | c | p{5cm} | p{5cm} |  p{5cm} | }
     \hline
      Dim & kNN Accuracy vs k & 4-fold cross-validation Accuracy & Confusion Matrix \\ \hline
      2-D
      &
      
     \includegraphics[width=0.3\textwidth, height=40mm]{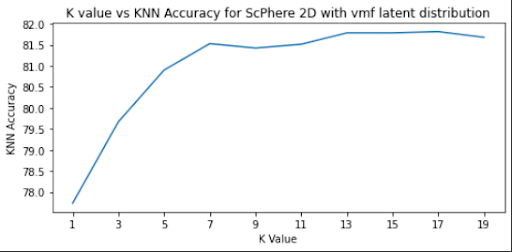}
      & 
    \includegraphics[width=0.3\textwidth, height=40mm]{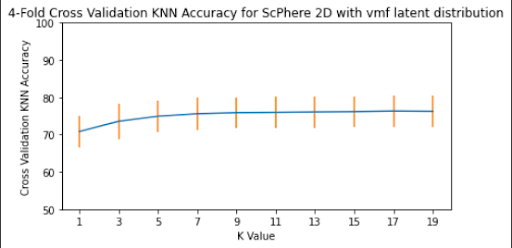}
      & 
    \includegraphics[width=0.3\textwidth, height=40mm]{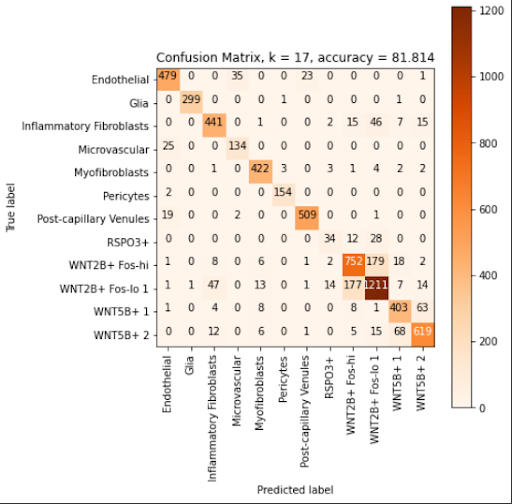}  
      \\ \hline
    10-D
      &
      
     \includegraphics[width=0.3\textwidth, height=40mm]{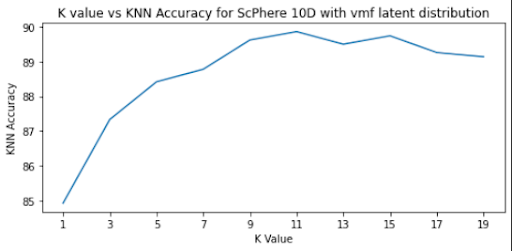}
      & 
    \includegraphics[width=0.3\textwidth, height=40mm]{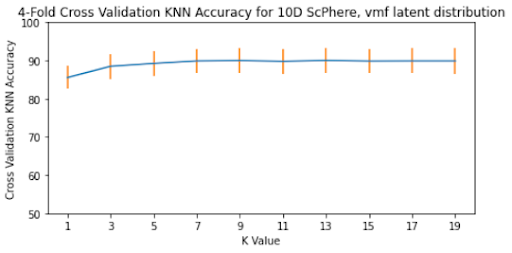}
      & 
    \includegraphics[width=0.3\textwidth, height=40mm]{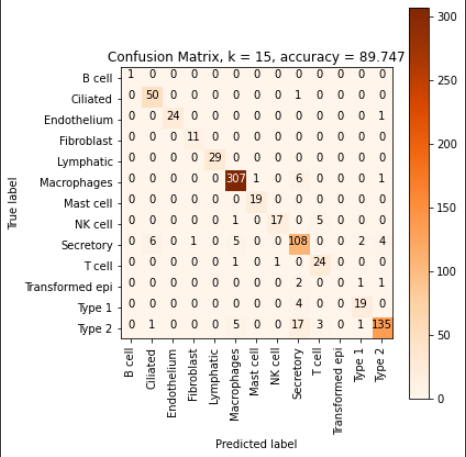}  
    \\ \hline
    20-D
      &
      
     \includegraphics[width=0.3\textwidth, height=40mm]{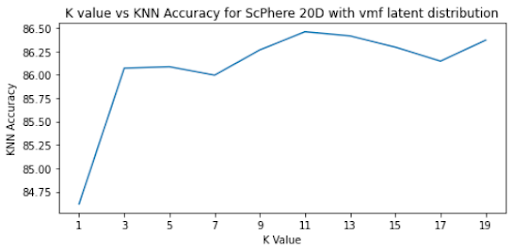}
      & 
    \includegraphics[width=0.3\textwidth, height=40mm]{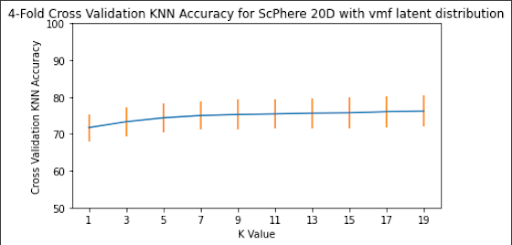}
      & 
    \includegraphics[width=0.3\textwidth, height=40mm]{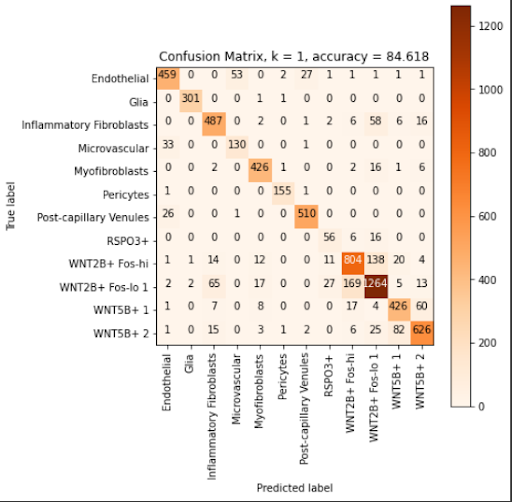}  
    \\ \hline
    
      \end{tabular}
      \label{tbl:myLboro}
      \end{center}
      \end{table}

\newpage
\clearpage
\subsection{scDHA Supplemental Figures} \label{scDHA_sup}
\begin{table}[!h]
     \begin{center}
    \caption{scDHA Results for 15D for All Datasets}
     \begin{tabular}{ | >{\centering\arraybackslash}m{1cm} | p{7.5cm} | p{7.5cm} | }
     \hline
      Data & Latent Distribution & 4-fold cross-validation Accuracy\\ \hline
      A
      &
      
     \includegraphics[width=0.45\textwidth, height=50mm]{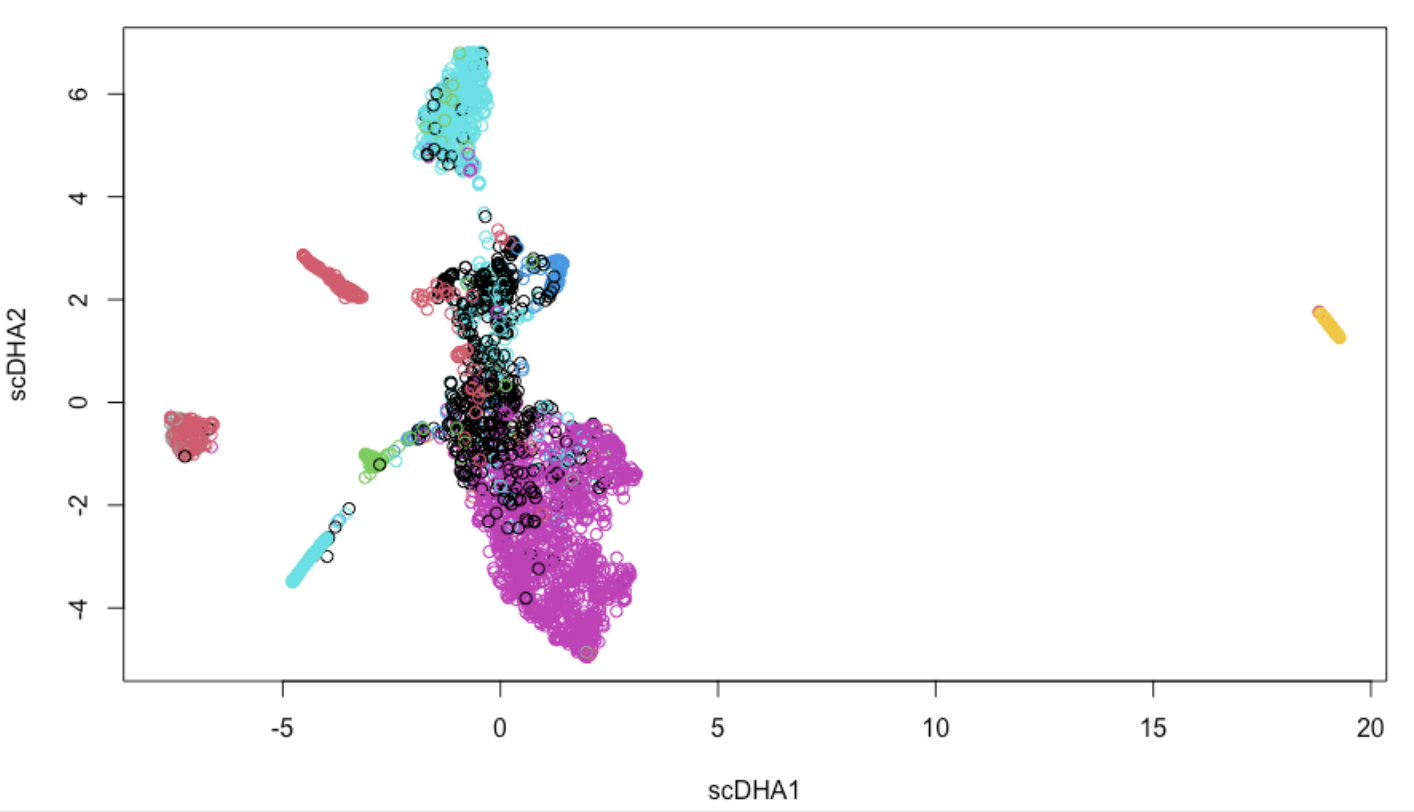}
      & 
    \includegraphics[width=0.45\textwidth, height=60mm]{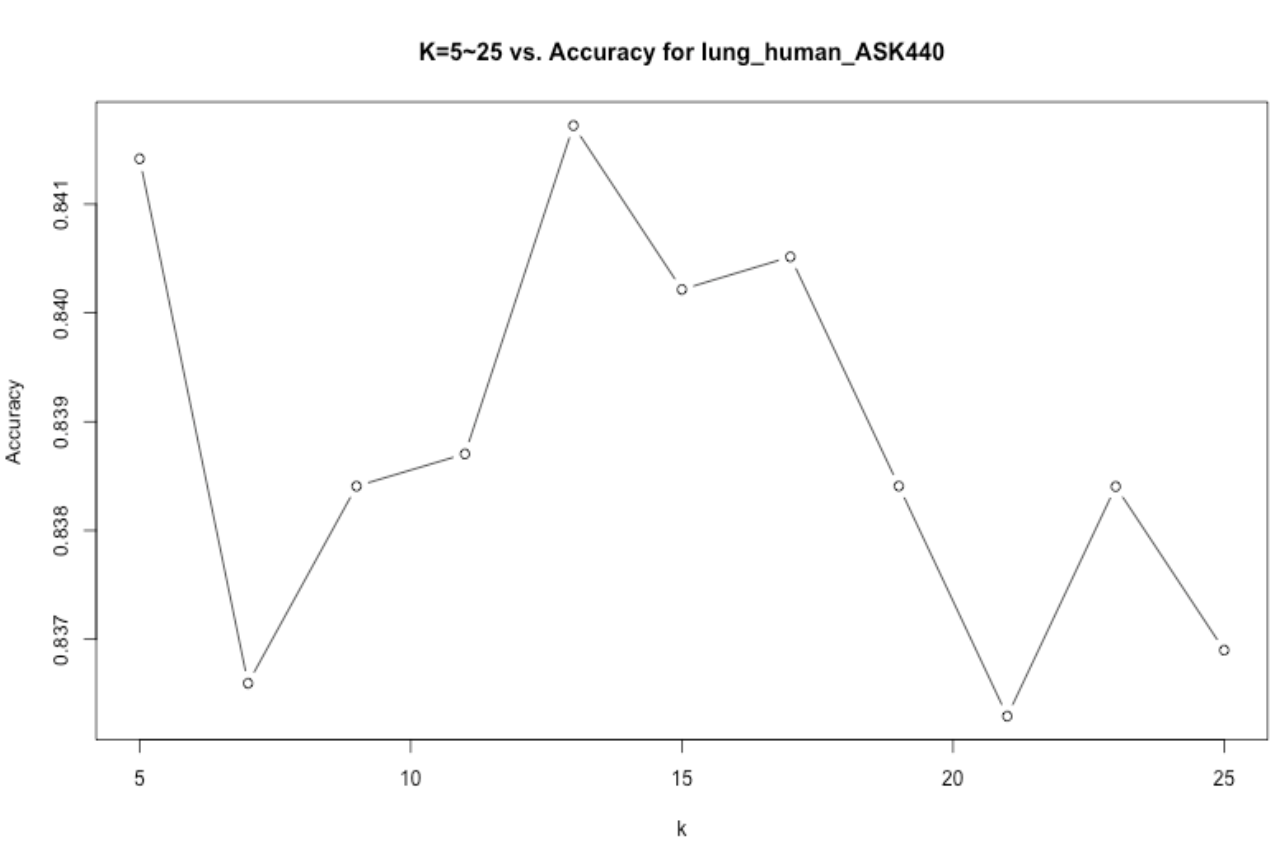}
      \\ \hline
    B
      &
      
     \includegraphics[width=0.45\textwidth, height=50mm]{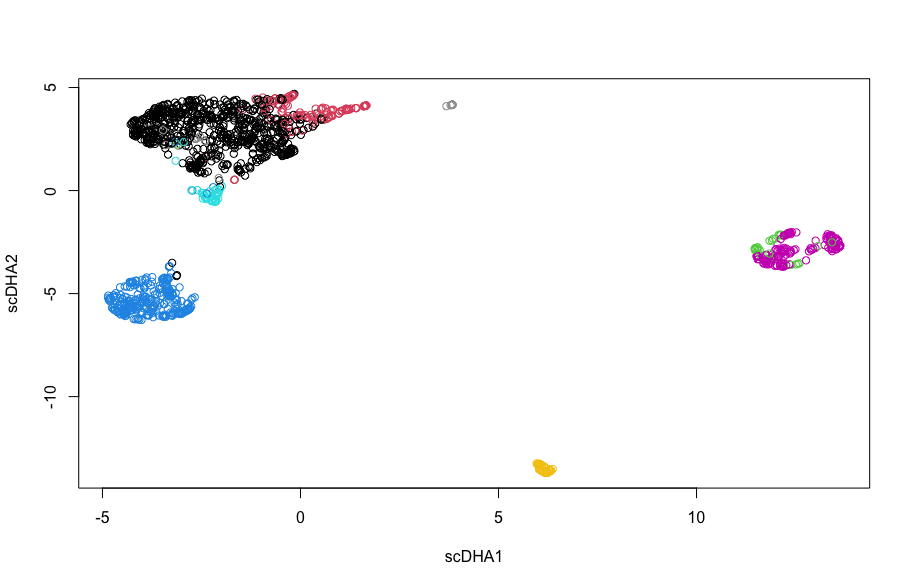}
      & 
    \includegraphics[width=0.45\textwidth, height=60mm]{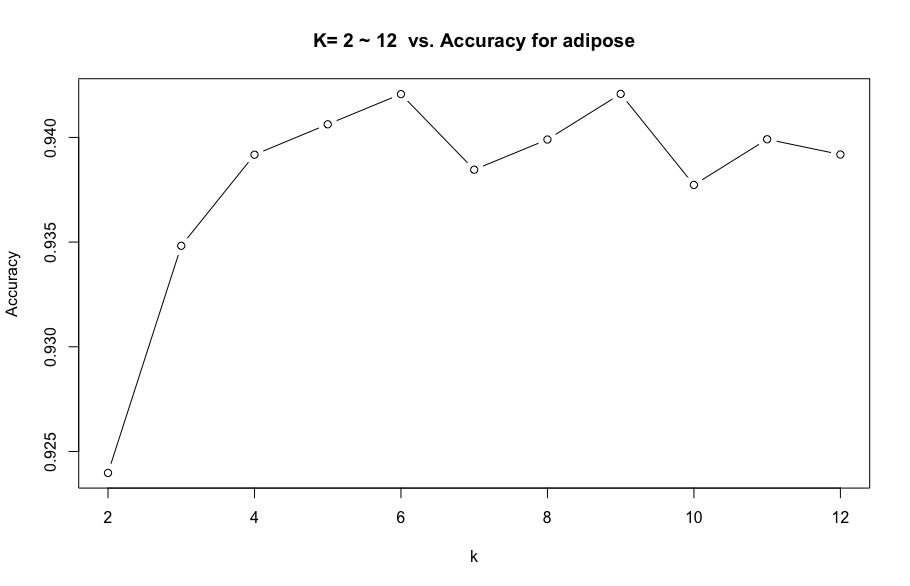}  
    \\ \hline
    C
      &
      
     \includegraphics[width=0.45\textwidth, height=50mm]{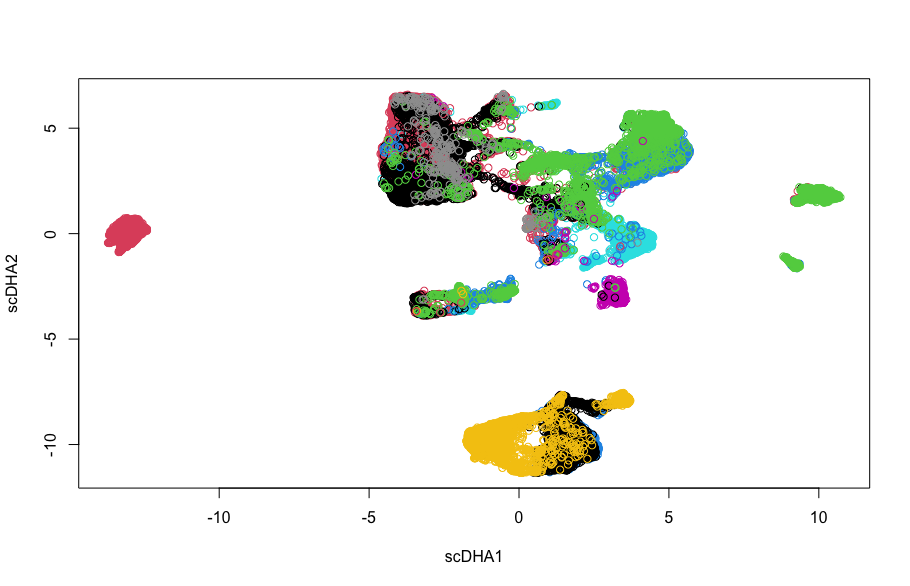}
      & 
    \includegraphics[width=0.45\textwidth, height=60mm]{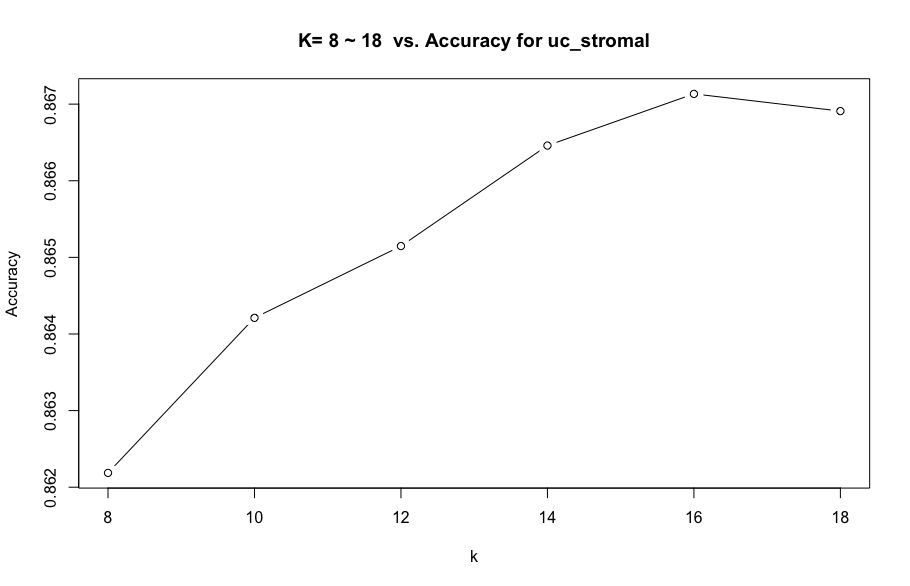} 
    \\ \hline
    
      \end{tabular}
      \label{tbl:myLboro}
      \end{center}
      \end{table}

\begin{figure}[!t]
    \centering
    \includegraphics[scale=0.6]{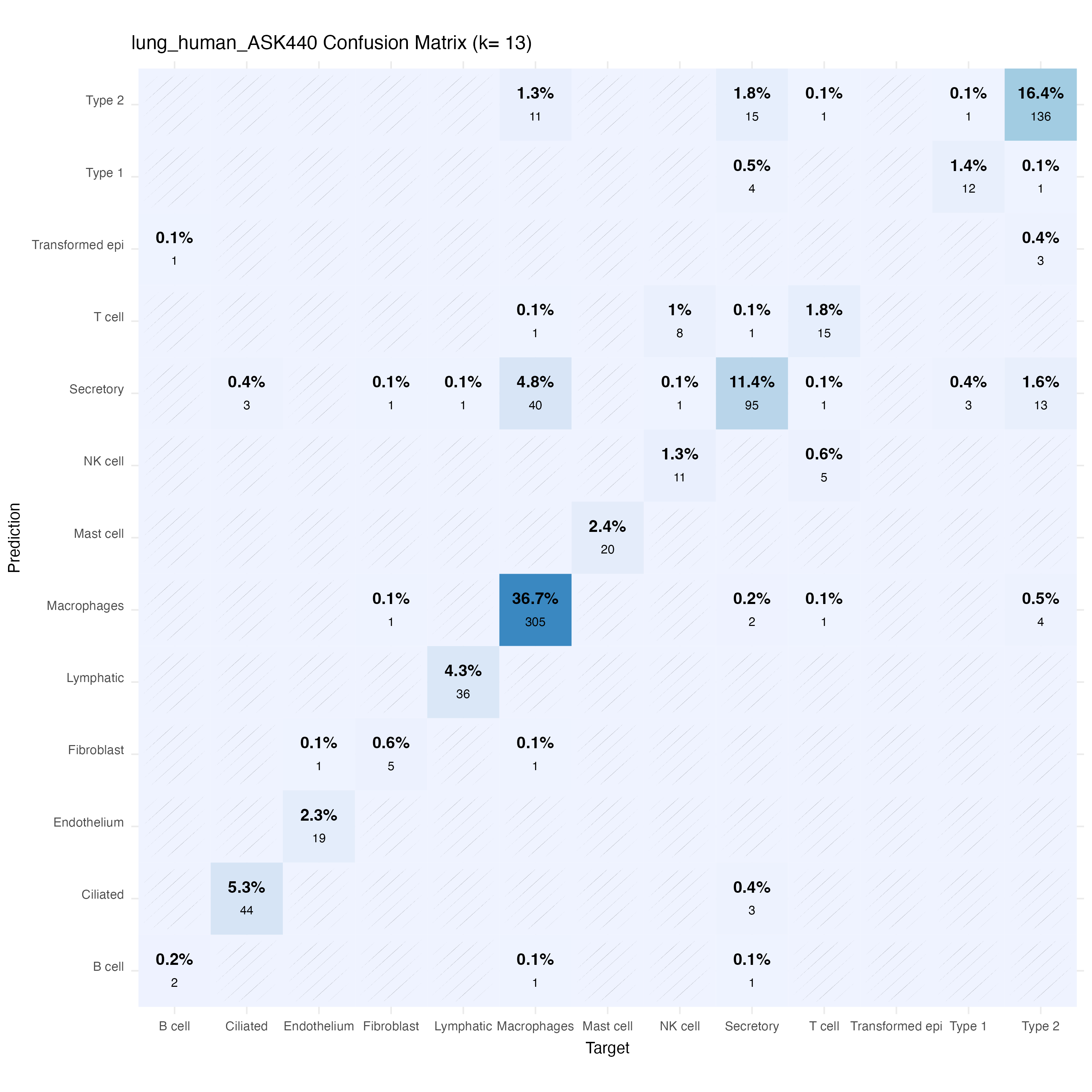}
    \caption{Confusion Matrix for Max kNN ($k = 13$)}
\end{figure}

\begin{figure}[!t]
    \centering
    \includegraphics[scale=0.6]{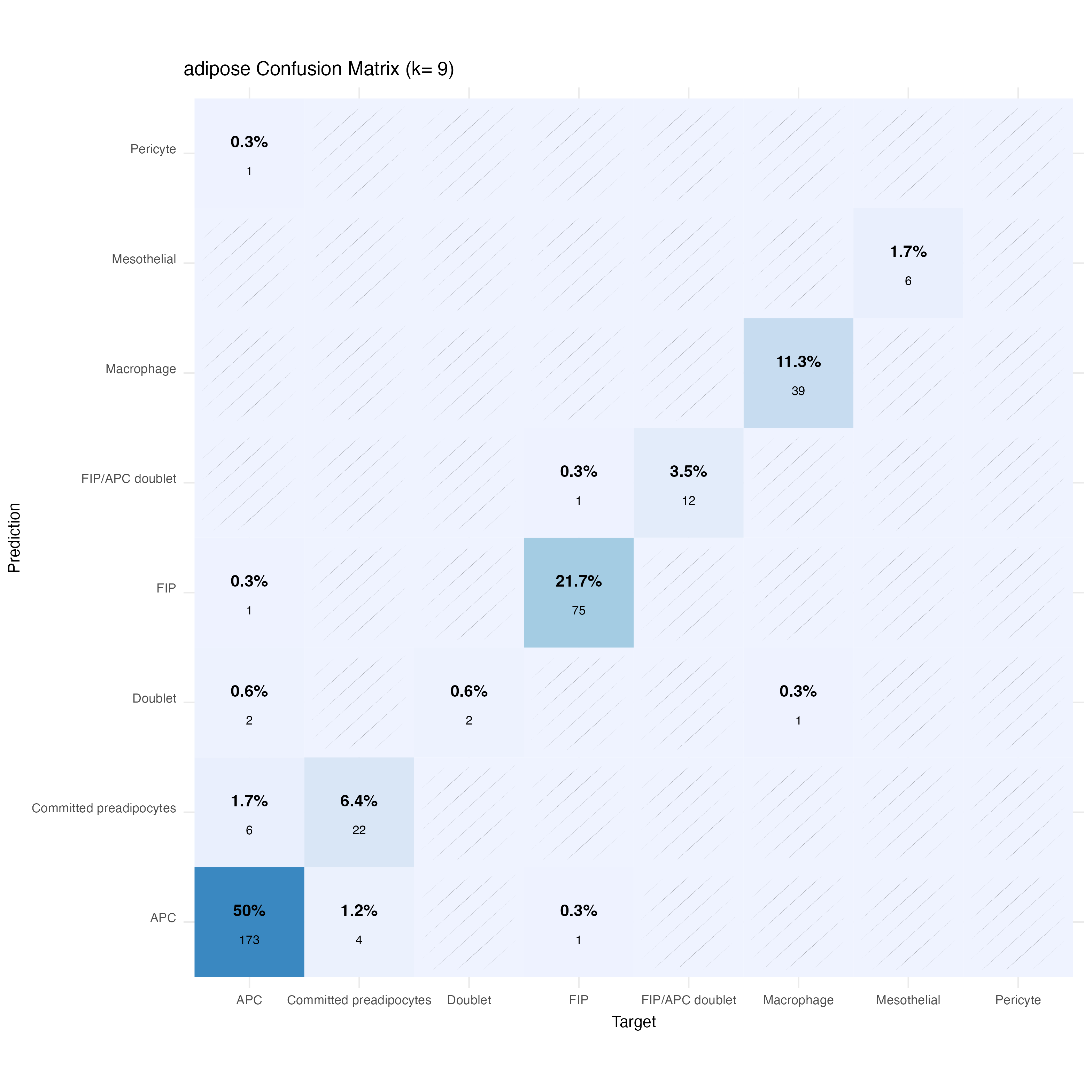}
    \caption{Confusion Matrix for Max kNN ($k = 9$)}
\end{figure}

\begin{figure}[!t]
    \centering
    \includegraphics[scale=0.6]{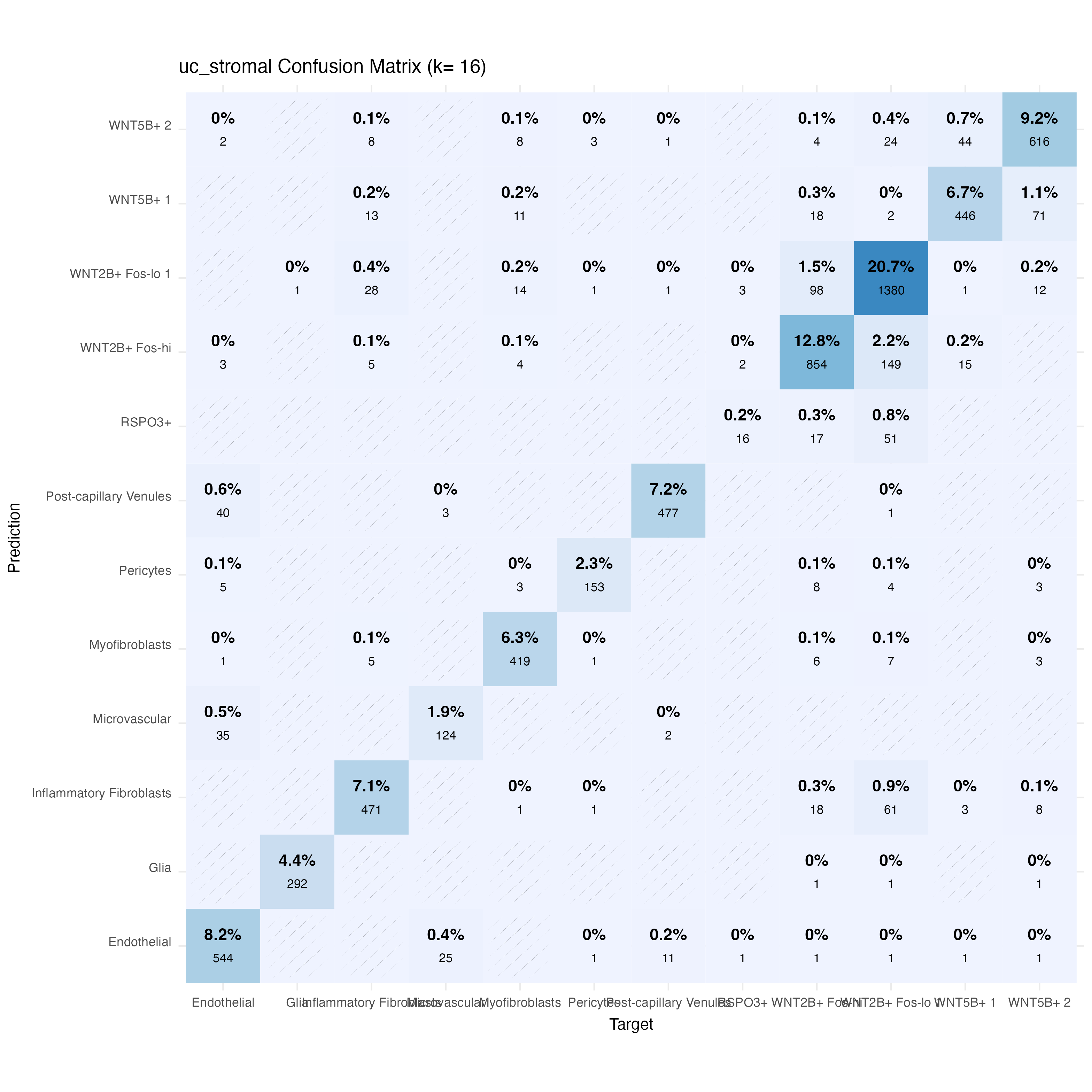}
    \caption{Confusion Matrix for kNN at max accuracy ($k = 16$)}
\end{figure}

\clearpage
\subsection{VASC Supplemental Figures} \label{VASC_sup}
\begin{table}[htb]
     \begin{center}
    \caption{VASC Results for 2D, 10D, 20D for Dataset A}
     \begin{tabular}{ | c | p{5cm} | p{5cm} |  p{5cm} | }
     \hline
      Dim & kNN Accuracy vs k & 4-fold cross-validation Accuracy & Confusion Matrix \\ \hline
      2-D
      &
      
     \includegraphics[width=0.3\textwidth, height=40mm]{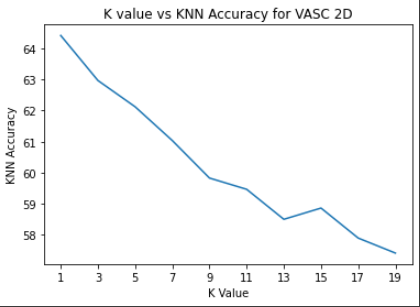}
      & 
    \includegraphics[width=0.3\textwidth, height=40mm]{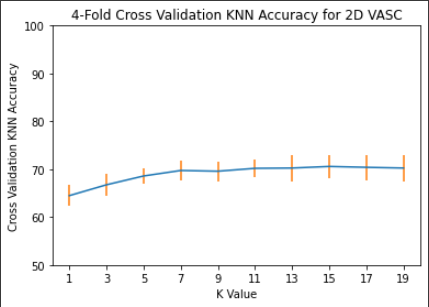}
      & 
    \includegraphics[width=0.3\textwidth, height=40mm]{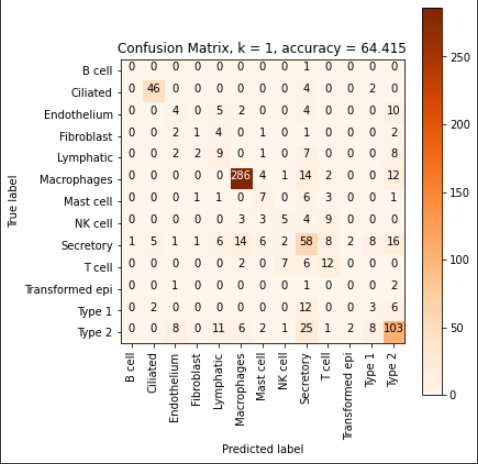}  
      \\ \hline
    10-D
      &
      
     \includegraphics[width=0.3\textwidth, height=40mm]{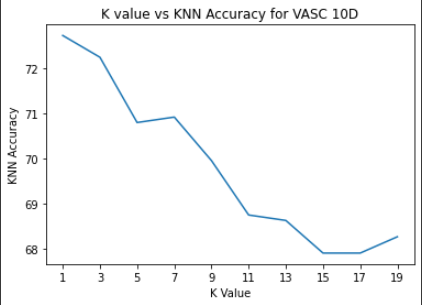}
      & 
    \includegraphics[width=0.3\textwidth, height=40mm]{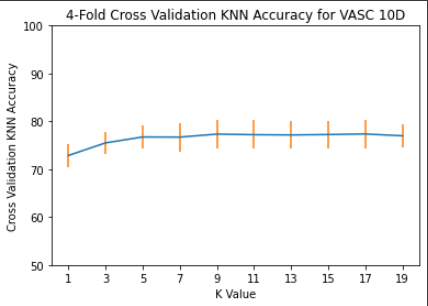}
      & 
    \includegraphics[width=0.3\textwidth, height=40mm]{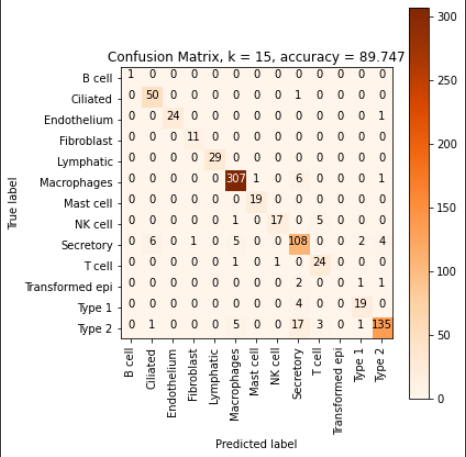}  
    \\ \hline
    20-D
      &
      
     \includegraphics[width=0.3\textwidth, height=40mm]{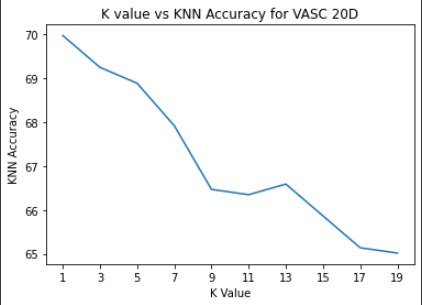}
      & 
    \includegraphics[width=0.3\textwidth, height=40mm]{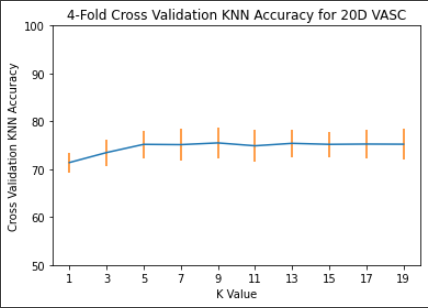}
      & 
    \includegraphics[width=0.3\textwidth, height=40mm]{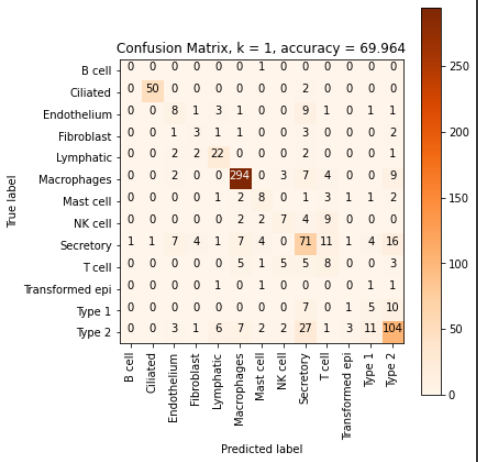}  
    \\ \hline
    
      \end{tabular}
      \label{tbl:myLboro}
      \end{center}
      \end{table}

\newpage

\begin{table}[htb]
     \begin{center}
    \caption{VASC Results for 2D, 10D, 20D for Dataset B}
     \begin{tabular}{ | c | p{5cm} | p{5cm} |  p{5cm} | }
     \hline
      Dim & kNN Accuracy vs k & 4-fold cross-validation Accuracy & Confusion Matrix \\ \hline
      2-D
      &
      
     \includegraphics[width=0.3\textwidth, height=40mm]{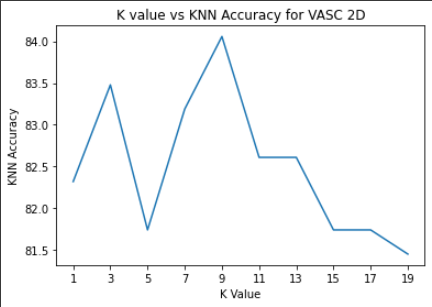}
      & 
    \includegraphics[width=0.3\textwidth, height=40mm]{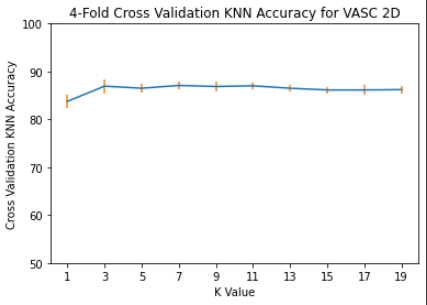}
      & 
    \includegraphics[width=0.3\textwidth, height=40mm]{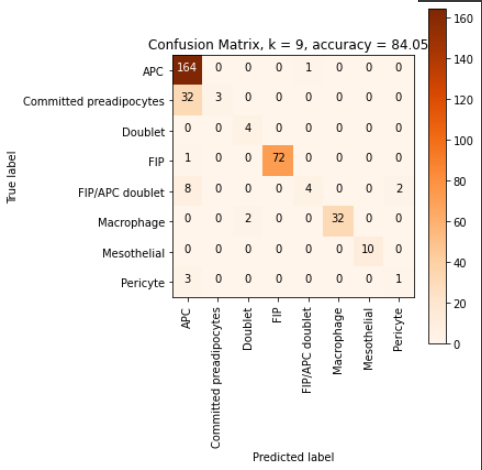}  
      \\ \hline
    10-D
      &
      
     \includegraphics[width=0.3\textwidth, height=40mm]{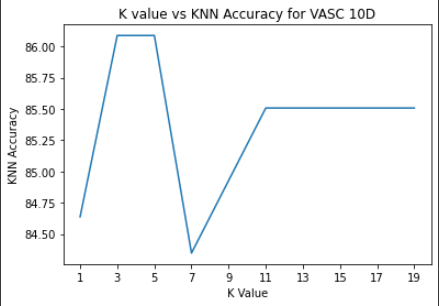}
      & 
    \includegraphics[width=0.3\textwidth, height=40mm]{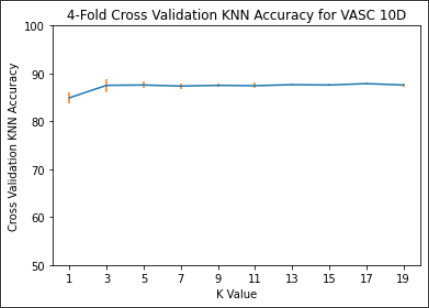}
      & 
    \includegraphics[width=0.3\textwidth, height=40mm]{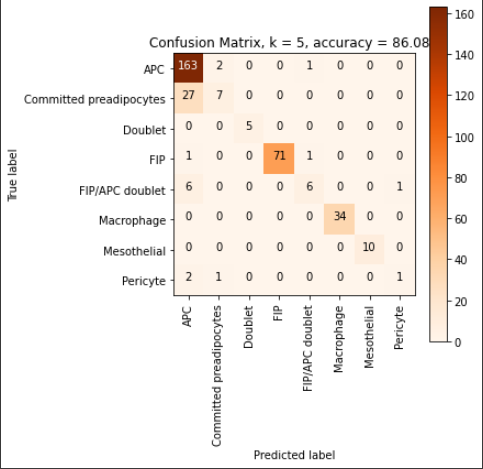}  
    \\ \hline
    20-D
      &
      
     \includegraphics[width=0.3\textwidth, height=40mm]{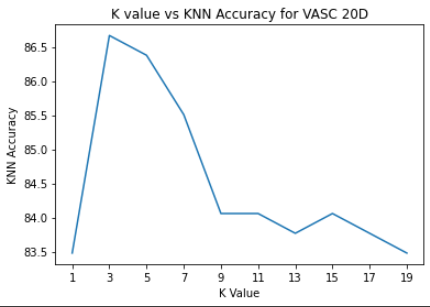}
      & 
    \includegraphics[width=0.3\textwidth, height=40mm]{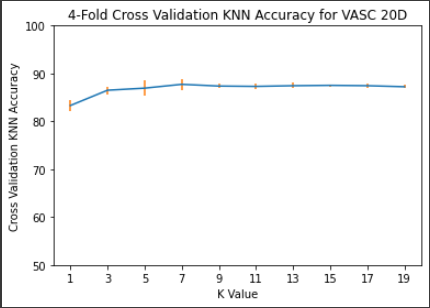}
      & 
    \includegraphics[width=0.3\textwidth, height=40mm]{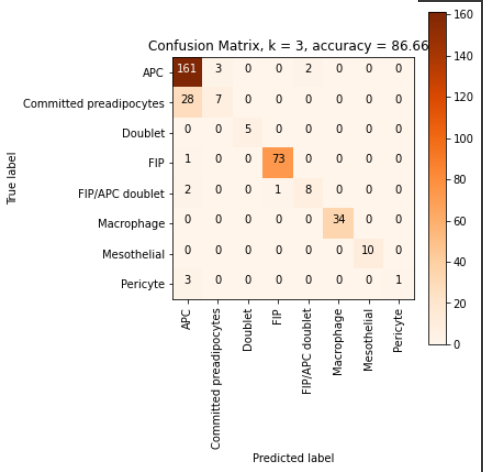}  
    \\ \hline
    
      \end{tabular}
      \label{tbl:myLboro}
      \end{center}
      \end{table}

\newpage
\begin{table}[htb]
     \begin{center}
    \caption{VASC Results for 2D, 10D, 20D for Dataset C}
     \begin{tabular}{ | c | p{5cm} | p{5cm} |}
     \hline
      Dim & kNN Accuracy vs k & 4-fold cross-validation Accuracy \\ \hline
      2-D
      &
      
     \includegraphics[width=0.3\textwidth, height=40mm]{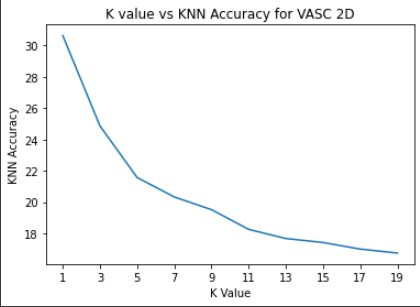}
      & 
    \includegraphics[width=0.3\textwidth, height=40mm]{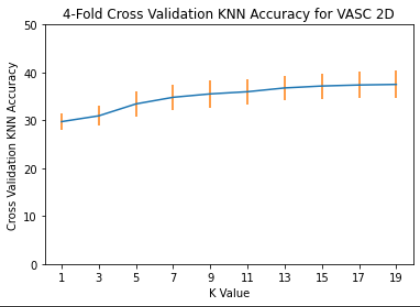}
      \\ \hline
    10-D
      &
     \includegraphics[width=0.3\textwidth, height=40mm]{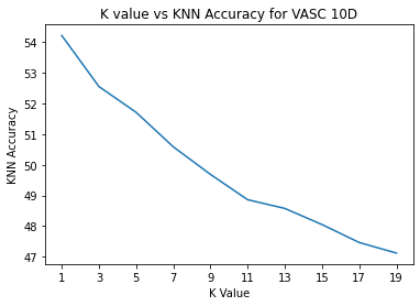}
      & 
    \includegraphics[width=0.3\textwidth, height=40mm]{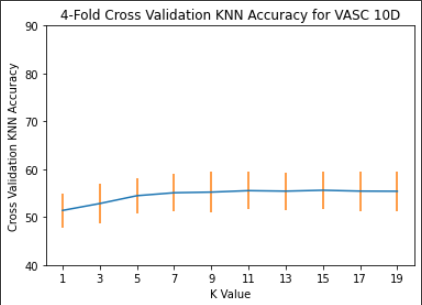}
    \\ \hline
    20-D
      &
      
     \includegraphics[width=0.3\textwidth, height=40mm]{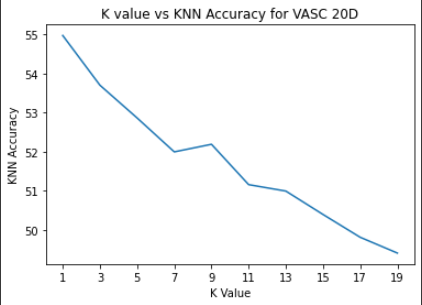}
      & 
    \includegraphics[width=0.3\textwidth, height=40mm]{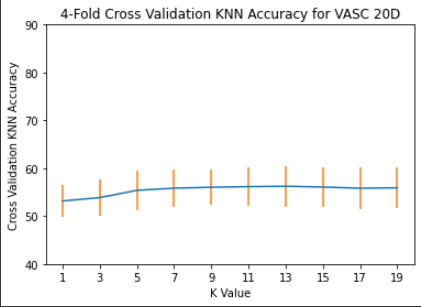}
    \\ \hline
    
      \end{tabular}
      \label{tbl:myLboro}
      \end{center}
      \end{table}
      
\newpage      

% The content of the Appendix begins here.
% Use the \verb|\section| command to create sections.

% Appendices are optional. An example of a single appendix displays in the template by default. Follow the instructions embedded in the template code for multiple appendices. Use the \verb|\ref{ }| command to reference appendices in your text.

\end{appendices}

%-------------------------------------------
% To insert multiple appendices that are sequentially lettered (Appendix A, Appendix B, etc.), uncomment lines 252-260. If you do, you will also need to comment out lines 243-249.
%\begin{appendices}
%\multappendices
%\chpt{Title of Appendix A}
%The content of Appendix A begins here. Use the \verb|\chpt| command to insert additional appendices.
%\section{Section Name of Appendix A}
%Use the \verb|\section| command to create sections.
%\chpt{Title of Appendix B}
%The content of Appendix B begins here.
%\end{appendices}
%-------------------------------------------

%%%%% BIBLIOGRAPHY %%%%%
%-------------------------------------------
\begin{bibliof}
%\nocite{*} % If applicable, uncomment this line to display all entries in the .bib file
\bibliography{bibliography}
\end{bibliof}
%-------------------------------------------
\end{document}